\documentclass{article}

\usepackage{microtype}
\usepackage{graphicx}
\usepackage{subcaption}
\usepackage{listings}
\usepackage{xcolor}
\usepackage{booktabs} 

\definecolor{keyword}{rgb}{0.26, 0.44, 0.76}
\definecolor{comment}{rgb}{0.5, 0.5, 0.5}
\definecolor{string}{rgb}{0.56, 0.93, 0.56}
\definecolor{backcolour}{rgb}{0.95, 0.95, 0.92}

\lstdefinestyle{pseudocode}{
    language=Python,
    basicstyle=\ttfamily\footnotesize,
    keywordstyle=\color{keyword}\bfseries,
    commentstyle=\color{comment}\itshape,
    stringstyle=\color{string},
    backgroundcolor=\color{backcolour},
    showstringspaces=false,
    breaklines=true,
    frame=single,
    captionpos=b,
    morekeywords={input, output, while, for, if, else, return}  
}
\usepackage{hyperref}

\usepackage[accepted]{icml2025}

\usepackage{amsmath}
\usepackage{amssymb}
\usepackage{mathtools}
\usepackage{amsthm}

\usepackage[capitalize,noabbrev]{cleveref}

\theoremstyle{plain}

\theoremstyle{definition}

\theoremstyle{remark}

\usepackage[textsize=tiny]{todonotes}

\icmltitlerunning{Deep Koopman-based causal discovery}

\begin{document}

\twocolumn[
\icmltitle{Deep Koopman operator framework for causal discovery \\
in nonlinear dynamical systems}



\icmlsetsymbol{equal}{*}

\begin{icmlauthorlist}
\icmlauthor{Juan Nathaniel}{equal,cu,leap}
\icmlauthor{Carla Roesch}{equal,cu,leap}
\icmlauthor{Jatan Buch}{cu,leap}
\icmlauthor{Derek DeSantis}{lanl}
\icmlauthor{Adam Rupe}{pnnl}
\icmlauthor{Kara Lamb}{cu,leap}
\icmlauthor{Pierre Gentine}{cu,leap}
\end{icmlauthorlist}

\icmlaffiliation{cu}{Columbia University}
\icmlaffiliation{pnnl}{Pacific Northwest National Laboratory}
\icmlaffiliation{lanl}{Los Alamos National Laboratory}
\icmlaffiliation{leap}{LEAP NSF Science and Technology Center}

\icmlcorrespondingauthor{Juan Nathaniel}{jn2808@columbia.edu}
\icmlcorrespondingauthor{Carla Roesch}{cmr2293@columbia.edu}

\icmlkeywords{Machine Learning, ICML}

\vskip 0.3in
]



\printAffiliationsAndNotice{\icmlEqualContribution} 

\begin{abstract} We use a deep \textbf{K}oopman operator-theoretic formalism to develop a novel c\textbf{ausal} discovery algorithm, \texttt{Kausal}. Causal discovery aims to identify cause-effect mechanisms for better scientific understanding, explainable decision-making, and more accurate modeling. Standard statistical frameworks, such as Granger causality, lack the ability to quantify causal relationships in nonlinear dynamics due to the presence of complex feedback mechanisms, timescale mixing, and nonstationarity. This presents a challenge in studying many real-world systems, such as the Earth's climate. Meanwhile, Koopman operator methods have emerged as a promising tool for approximating nonlinear dynamics in a linear space of observables. In \texttt{Kausal}, we propose to leverage this powerful idea for causal analysis where optimal observables are inferred using deep learning. Causal estimates are then evaluated in a reproducing kernel Hilbert space, and defined as the distance between the marginal dynamics of the effect and the joint dynamics of the cause-effect observables. Our numerical experiments demonstrate \texttt{Kausal}'s superior ability in discovering and characterizing causal signals compared to existing approaches of prescribed observables. Lastly, we extend our analysis to observations of El Ni\~no-Southern Oscillation highlighting our algorithm's applicability to real-world phenomena. Our code is available at \url{https://github.com/juannat7/kausal}.
\end{abstract}

\section{Introduction}\label{introduction}

Causal discovery seeks to disentangle the underlying cause-effect mechanisms of physical processes, which is crucial for improved scientific understanding \cite{aloisi2022effect,cattry2025ecopro,roesch2025decreasing,williams2025precipitation}. Since performing interventions to detect causality is often infeasible in many domains, there is a great interest to develop data-driven causal discovery methods \citep[e.g.,][]{granger1969investigating,hyvarinen2010estimation,runge2020discovering}. However, they tend to operate under certain assumptions, such as time-invariance, stationarity, and linearity~\cite{granger1969investigating,runge2019inferring,bareinboim2022pearl, camps-vallsDiscoveringCausalRelations2023}, which limit their applicability.\footnote{Causal discovery is different from causal effect estimation, where the latter aims to quantify the causal effect between variables for an existing (either discovered or assumed) causal graph.} For instance, these simplifications fail while examining nonlinear systems due to the presence of complex feedback mechanisms, spatial autocorrelation, and timescale mixing \cite{lorenz1963deterministic,rupe2024principles}. 

In general, there are two categories of approaches for inferring causal relationships in nonlinear dynamical systems: i) using information flow to identify causal influence among variables and subspaces~\cite{granger1969investigating,liangInformationTransferDynamical2005, majdaInformationFlowSubspaces2007, jamesInformationFlowsCritique2016,tank2021neural}; and ii) performing a lifting transformation to project the system of interest to a higher-dimensional space where causal effects can be investigated more easily~\cite{lorenz1991dimension, sugiharaDetectingCausalityComplex2012, gilpinGenerativeLearningNonlinear2024,butler2024tangent}. Thus, we pose our research question: Can we combine these two principles to apply linear causal analysis methods in nonlinear systems? 

A natural foundation for us to start answering this question is Koopman theory \cite{koopman1931hamiltonian, mezicAnalysisFluidFlows2013, bruntonModernKoopmanTheory2022}, which states that finite nonlinear dynamics can be represented with an infinite-dimensional linear operator acting on the space of all possible measurement functions. This formalism presents a significant opportunity for the application of standard linear methods for system analysis, estimation, and control \cite{bruntonMachineLearningSparse2025}. Recent work by \citet{rupeCausalDiscoveryNonlinear2024} has established a theoretical foundation linking Koopman theory to causal analysis for dynamical systems using prescribed embeddings. Simultaneously, efficiently and effectively estimating optimal embeddings has been enabled by rapid advances of deep learning in high-dimensional settings~\cite{bruntonKoopmanInvariantSubspaces2016,luschDeepLearningUniversal2018,wang2022koopman}. Therefore, data-driven Koopman operator methods have seen a surge in interest, especially for spatiotemporal analysis in a number of fields~\cite{nathan2018applied, azencotForecastingSequentialData2020a,riceAnalyzingKoopmanApproaches2021,froylandSpectralAnalysisClimate2021,lintnerIdentificationMaddenJulian2023,lambReducedOrderModelingLinearized2024,nathaniel2025generative}. 

\begin{figure*}[h!]
    \centering
    \includegraphics[width=0.75\linewidth]{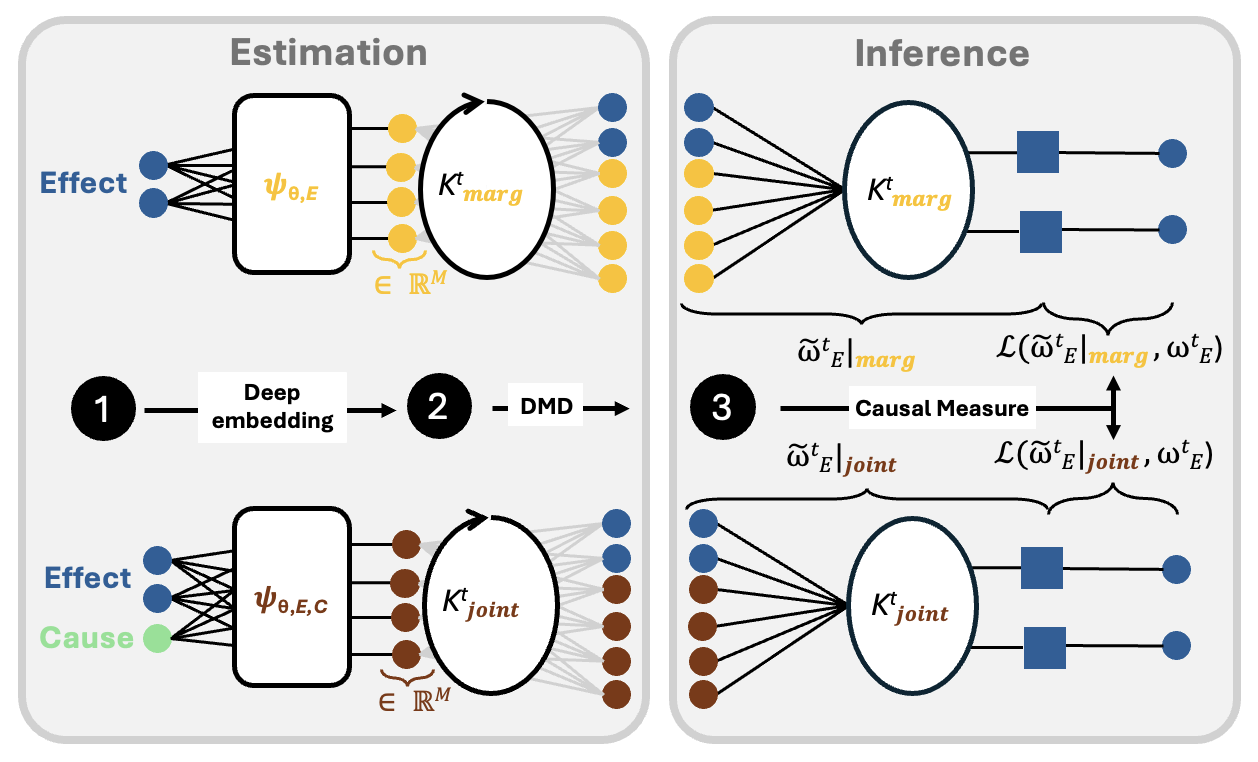}
    \caption{Schematic of the \texttt{Kausal} algorithm. We 1) estimate the embeddings using deep learning, and 2) approximate the Koopman operator with dynamic mode decomposition (DMD). Then, 3) we infer causal measures by computing the difference in prediction error between marginal (effect-only) and joint (effect-cause) models.}
    \label{fig:algorithm-schematic}
\end{figure*}

\textbf{Main contributions.} In this paper, we leverage ideas from deep learning, Koopman theory, and causal inference. Our main contributions can be summarized as follows:
\begin{itemize}
    \item We introduce a novel deep \textbf{K}oopman operator-based c\textbf{ausal} discovery framework, \texttt{Kausal}, as illustrated in Figure~\ref{fig:algorithm-schematic}.
    \item We show that \texttt{Kausal} infers a better representation of the basis functions through deep learning, as opposed to existing prescribed approaches, which leads to a more accurate and meaningful causal analysis in high-dimensional nonlinear systems. 
    \item We successfully apply \texttt{Kausal} to both idealized models and real-world phenomena highlighting the algorithm's scalability and generalizability.
\end{itemize}

\section{Methodology}
\subsection{Background}
In this section, we follow \citet{rupeCausalDiscoveryNonlinear2024} to briefly introduce the mathematical background and the necessary notation to introduce our novel causal algorithm in Section~\ref{sec:causality}.

\textbf{Dynamical systems.} Dynamical systems theory describes the evolution of a system over time, with the system state represented as a point in a phase space, $\omega \in \Omega$, which we assume to be in the Euclidean space $\mathbb{R}^N$. Throughout this paper variables are treated as vectors unless explicitly stated otherwise.

\textbf{Flow maps.} Time-evolution of states is defined through a set of functions called \textit{flow maps} $\{ 
\Phi^t : \Omega \rightarrow \Omega \}_{t \in T}$ that carries the phase space back onto itself. For an initial state $\omega \in \Omega$, flow maps evolve $\omega$ through time \textit{t}, given by ${\omega(t) = \Phi^t(\omega)}$. The \textit{orbit} of a state $\omega$ is the set $\{ \omega (t) = \Phi^t(\omega)\}_{t \in T}$, representing the trajectory of the system over the time domain $T$. Although we denote the initial state as $\omega$, writing $\omega (t, \omega_0) = \Phi^t(\omega_0)$ with the initial state  $\omega_0$ would be equivalent. For simplicity, we also restrict our analysis to {\it autonomous} dynamical systems characterized by time-independent flow maps.

\textbf{Ordinary differential equations.} In a deterministic setting, the evolution of $\omega(t)$ over time $t$ can be described through \textit{ordinary differential equations} (ODEs) of the form:
\begin{equation}\label{eq:ode}
    \frac{d\omega}{dt} = f(t,\omega)
\end{equation}
where $f$ is an arbitrary function and the solution $\omega(t)$ depends on the initial condition $\omega(t_0) = \omega_0$ at time $t_0$.

\textbf{Components.} To formalize the idea of causality in a dynamical system, we need to define different \textit{components} of the system which can impact each other. For an $N$-dimensional phase space (i.e. $N$ degrees of freedom) as described above, we partition the space into disjoint system component subspaces as $\Omega = \Omega_1 \times \Omega_2 \times \dots \times \Omega_L$ with $\omega_i \in  \Omega_i$ and $\omega = [\omega_1 \text{ } \omega_2 \text{ } \dots \text{ } \omega_L]$, where $L \leq N$.

\textbf{Reproducing Kernel Hilbert Spaces (RKHS).} Koopman operators (described in the next section) act on functions rather than directly on system states, mapping one function to another by evolving it under the system dynamics. To ensure the operations of the Koopman operator are mathematically consistent and computationally feasible, we therefore require a well-defined function space. For data-driven analysis, a natural choice of function spaces is reproducing kernel Hilbert spaces.

A RKHS is a Hilbert space of functions $\mathcal{F}$ such that the evaluation $f \mapsto f(\omega)$ is continuous for each point $\omega \in \Omega$.  This requires the existence of a kernel function $k:\Omega \times \Omega \rightarrow \mathbb{C}$ such that $k(\cdot, \omega') \in \mathcal{F}$ and $f(\omega') = \langle f, k(\cdot, \omega') \rangle$. The Moore-Aronszajn theorem~\cite{aronszajn1950theory} implies that a positive definite kernel function defines a unique RKHS, so we use the notation $\mathcal{F} = \mathcal{H}(k)$ unambiguously. We will express the kernel $k$ in terms of an associated feature map $\boldsymbol{\psi}: \Omega \rightarrow \mathcal{F}$, namely $k(\omega, \omega') = \langle \boldsymbol{\psi}(\omega), \boldsymbol{\psi}(\omega')\rangle_\mathcal{H}$.
The kernel $k$ is constructed as the tensor product of the kernels $k_i$, which are the kernel functions for each component $\Omega_i$ of the partitioned phase space $\Omega = \Omega_1 \times \cdots \times \Omega_L$. For a subspace $\Omega_X$ given $X \subset \{1,2,...,L \}$ we define the component observable function subspace $\mathcal{F}_X$ which is itself a RKHS. Thus, $\mathcal{F}_X = \mathcal{H}(k_X)$ is given through the kernel:
\begin{equation}
\begin{aligned}
    k_{X}(\omega, \omega') &:= \prod_{j \in X} k_j(\boldsymbol{\psi}(\omega_j), \boldsymbol{\psi}(\omega'_j)) \\ 
    &:= \langle \boldsymbol{\psi}(\omega_j), \boldsymbol{\psi}(\omega'_j) \rangle_\mathcal{H}
\end{aligned}
\end{equation}
for all $\omega, \omega' \in \Omega$. We summarize key features of RKHS relevant to the Koopman framework in Table~\ref{tab:rkhs_aspects}.

\subsection{Koopman theory}
\textbf{Koopman operator.} Koopman theory provides a \textit{global linearization} of high-dimensional nonlinear dynamical systems \cite{koopman1931hamiltonian, bruntonModernKoopmanTheory2022}. Specifically, the nonlinear temporal evolution of system states is represented as a linear flow in the evolution of system observables. The map from system states to observables is given by a function, $\boldsymbol{\psi}: \Omega \rightarrow \mathcal{F}$.

The \textit{Koopman operator} $\mathcal{K}^t: \mathcal{F} \rightarrow \mathcal{F}$ therefore acts as a time shift \textit{t} on an observable $\boldsymbol{\psi}(\omega)$:
\begin{equation}
    \boldsymbol{\psi}^t(\omega) = [\mathcal{K}^t\boldsymbol{\psi}](\omega) := [\boldsymbol{\psi} \circ \Phi^t](\omega) = \boldsymbol{\psi}(\omega(t)).
\end{equation}
We need Koopman operators acting on $\mathcal{F}$ to be bounded to ensure the continuity of the time-evolution of $\boldsymbol{\psi}$, i.e. $\mathcal{K}^t\boldsymbol{\psi}$. This property is given by defining the Koopman operator over a RKHS \cite{kostic2022learning,rupeCausalDiscoveryNonlinear2024}. 

\textbf{Observables.} Obtaining observables in which the nonlinear dynamics appear approximately linear is an open challenge \cite{bruntonModernKoopmanTheory2022}. Existing works often rely on a set of $M$-dimensional dictionary functions $\boldsymbol{\psi} = [ \psi_1 \, \psi_2  \, \cdots \, \psi_M ]^\intercal$ spanning a finite-dimensional Hilbert space $\mathcal{F}_{\boldsymbol{\psi}} \subset \mathcal{F}$, such as Gaussian, and Random Fourier features~\citep[RFF;][]{rahimi2007random}, Polynomial, Time Delay~\citep[TDF;][]{abarbanel1994predicting}. Note that these dictionary functions are the previously defined feature maps. Often these functions may not be sufficiently expressive to represent complex dynamics \cite{kostic2022learning}. Nonetheless, deep learning provides a natural framework for inferring the optimal high-dimensional embedding, which constitutes one of the goals of this paper. 

In this paper, we compute a finite-rank approximation of the Koopman operator with dictionary functions ${\boldsymbol{\psi}_\theta: \Omega \rightarrow \mathcal{H}}$ that are learned using different neural network-based encoder-decoder architectures. Since we use appropriate regularizations and activations, namely the bounded sigmoid function, we ensure that the observables are finite. As a result, the kernel function is also bounded, i.e., ${k_\theta(\omega, \omega') = \langle\boldsymbol{\psi_\theta}(\omega), \boldsymbol{\psi_\theta}(\omega')\rangle_\mathcal{H} < \infty}$ for all ${\omega, \omega' \in \Omega}$~\cite{aronszajn1950theory, caponnetto2007optimal, kostic2022learning}. The size of our encoder final layer's output vector $M \in \mathbb{Z}_+$ determines the dimensionality of $\mathcal{H}$, which is the number of learned dictionary functions. 

\textbf{Dynamic mode decomposition.} To estimate the Koopman operator $\mathcal{K}^t$ we use \textit{dynamic mode decomposition} (DMD), which solves a least-squares regression problem to identify a finite-dimensional linear representation of the system dynamics. Using dictionary functions defined as, 
\begin{equation}
    \boldsymbol{\psi_\theta}(\omega(n))) = \left[ \psi_{\theta,1}(\omega(n)) \dots \psi_{\theta,M}(\omega(n)) \right]^\intercal \in \mathbb{R}^{M \times 1},
\end{equation}
we construct data matrices consisting of observables for $D \in \mathbb{Z}_{+}$ pairs of data points $\{ (\omega(n),\omega^t(n) \}_D$, denoted as $\Psi_\theta$ and $\Psi_\theta^t$ (\textit{t} time shifted):

\begin{equation}
    \Psi_\theta = \begin{bmatrix}
                | &  & | \\
                \boldsymbol{\psi}_\theta(\omega(n_1)) & \dots & \boldsymbol{\psi}_\theta(\omega(n_D)) \\
                | &  & | \\
             \end{bmatrix},
\end{equation}
\begin{equation}
    \Psi_\theta^t = \begin{bmatrix}
                | & & | \\
                \boldsymbol{\psi}_\theta^t(\omega(n_1)) & \dots & \boldsymbol{\psi}_\theta^t(\omega(n_D)) \\
                | & & | \\
             \end{bmatrix}.
\end{equation}

Assuming there exists a unique linear operator ${K^t: \mathcal{F}_{\Psi_\theta} \rightarrow \mathcal{F}_{\Psi_\theta}}$ that satisfies ${\langle \boldsymbol{\psi}'_{\theta}, \mathcal{K}^t \boldsymbol{\psi}_{\theta} \rangle_\mathcal{H} = \langle \boldsymbol{\psi}'_{\theta}, K^t \boldsymbol{\psi}_{\theta} \rangle}_\mathcal{H}$, where $\boldsymbol{\psi}_\theta, \boldsymbol{\psi}'_\theta \in \boldsymbol{\Psi}_\theta$, we can define the least-squares regression problem as:

\begin{equation}
    \begin{aligned}
        &\sum_{n=1}^D \left\Vert [\mathcal{K}^t\Psi_\theta] (\omega(n)) - K^t\Psi_\theta(\omega(n))\right\Vert^2 \\
        &= \sum_{n=1}^D \left\Vert [\Psi_\theta \circ \Phi^t] (\omega(n)) - K^t\Psi_\theta(\omega(n))\right\Vert^2 \\
        &= \sum_{n=1}^D \left\Vert \Psi_\theta(\omega^t(n)) - K^t\Psi_\theta(\omega(n))\right\Vert^2 .
    \end{aligned}
\end{equation}
The full rank least-squares solution is given as:
\begin{equation}
    K^t := \Psi_\theta^t \Psi_\theta^\dagger    
\end{equation}
where $\Psi_\theta^\dagger$ is the pseudo-inverse of $\Psi_\theta$. A low-rank approximation is also feasible with matrix projection onto dominant modes using e.g., SVD. Note that $K^t$ is an approximation and converges to the Koopman operator $\mathcal{K}^t$ for $D,M \rightarrow \infty$ \cite{kordaConvergenceExtendedDynamic2018}. 

\section{Causal discovery}\label{sec:causality}
Here we connect principles of causal discovery to the deep Koopman framework introduced in the last section.

\subsection{Causal discovery in dynamical systems}

\textbf{Structural dynamical causal models.} In statistical terms, causal mechanisms are described through \textit{structural causal models} (SCM), where a system of $d$ random variables ${\boldsymbol{x} = \{x^1, \dots, x^d \}}$ are expressed as an arbitrary function $f^k$ of their direct parents (causes) $\boldsymbol{x}_{\text{PA}_k}$ and an exogenous distribution of noise $\epsilon^k$ \cite{peters2022causal}:
\begin{equation}\label{eq:scm}
    x_k := f^k(\boldsymbol{x}_{\text{PA}_k}, \epsilon^k) \text{, for } k = 1, \dots, d
\end{equation}

For dynamical systems, we can extend SCM for a collection of $d$ ODEs (Equation~\ref{eq:ode}) to define \textit{structural dynamical causal models} (SDCM) \cite{rubenstein2018deterministicodesdynamicstructural,peters2022causal,boeken2024dynamic}:
\begin{equation}
    \frac{\text{d}}{\text{d}t} \omega_{k,t} := f^k(\boldsymbol{\omega}_{{\text{PA}_k,t}}), \text{with } \omega_{k,0} = \omega_k(0)
\end{equation}
where $k \in \{1, \dots, d\}$. 

\begin{figure}[h]
  \centering
    \includegraphics[width=0.7\linewidth]{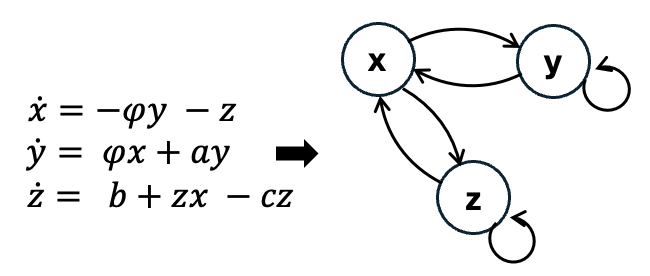}
    \caption{Illustration of ODEs describing the R\"{o}ssler Oscillator system (left), and the corresponding causal graph (right).}
    \label{fig:sdcm}
\end{figure}

\textbf{Causal graph.} The structural assignment of the SDCM induces a \textit{directed acyclic graph} (DAG) $\mathcal{G} = (\mathcal{V},\mathcal{E})$ over the variables $\omega_k$. $\mathcal{G}$ includes nodes $v_k \in \mathcal{V}$ for every $\omega_k \in \boldsymbol{\omega}$ and directed edges $(k,i) \in \mathcal{E}$ if $\omega_k \in \boldsymbol{\omega}_{\text{PA}_i}$. The corresponding DAG is visualized in Figure~\ref{fig:sdcm}. 

\textbf{Causal mechanism.} Using the previously defined framework of component subspaces for dynamical systems, we partition the phase space $\Omega$ of a system into components: 
\begin{equation}
    \Omega = \Omega_E \times \Omega_C \times \Omega_R
\end{equation} 
representing the effect, cause, and remainder components, respectively. Each state $\omega \in \Omega$ is projected into the effect component using coordinate projections \textit{$P_E$}, and the evolution of $\omega_E(t)$ is tracked through the \textit{component flow map}:
\begin{equation}
    P_E \Phi_t : \omega = [\omega_E \, \omega_C \, \omega_R]^\intercal \mapsto \omega_E(t) .
\end{equation}
This framework is flexible enough to allow for multipartite causal analysis, i.e., multiple causal factors $\Omega_j$ and $\Omega_k$, by combining them into a single component $\Omega_j \times \Omega_k$.

\textbf{Causal influence.} The existence of a dynamical causal relationship at time \textit{t} is determined by whether the state of the effect component depends on the initial state of the cause component, i.e., $\Omega_C$ \textit{dynamically causally influences} $\Omega_E$ at time \textit{t}. Mathematically we write this as:
\begin{subequations}
    \begin{align}
        \Omega_C \rightarrow^t \Omega_E &\Longleftrightarrow \omega_E(t) \text{ is influenced by } \omega_C \text{ and }  \\
        \Omega_C \not\rightarrow^t \Omega_E &\Longleftrightarrow \omega_E(t) \text{ is } \emph{not} \text{ influenced by } \omega_C , \label{eq-causal-influence}
    \end{align}
\end{subequations}
where Equation~\ref{eq-causal-influence} implies $\Omega_C$ does \textit{not dynamically causally influence} $\Omega_E$.

\subsection{Causal discovery with deep Koopman operators}
Estimating dynamical causality directly is challenging for nonlinear dynamical systems, requiring carefully crafted counterfactual analysis \cite{merlisInteractingComponentsTopofatmosphere2014}. 

\textbf{Koopman causal influence.} Fortunately, we can rephrase dynamical causality using the Koopman framework, where it is represented by the linear evolution of observables in function spaces. This formulation necessitates that the observables associated with individual components form closed subspaces in the full RKHS $\mathcal{F}$ that represents the joint dynamical system. We argue that causal relationships in the state space $\Omega$ also induce distinguishable patterns in the flow of functions $\boldsymbol{\psi_\theta} \in \mathcal{F}$. 

For instance, if $\Omega_C \rightarrow^t \Omega_E$, then the application of the Koopman operator on the effect observables $\mathcal{F}_E$ will inherit this dependence which is expressed as $K^t\boldsymbol{\psi_\theta} \in \mathcal{F}_{E,C,R}$. We define this as $\Omega_C$ \textit{Koopman causally influences} $\Omega_E$ which is denoted by $\Omega_C \rightarrow^t_K \Omega_E$. Correspondingly, if $\Omega_C$ \textit{does not Koopman causally influence} $\Omega_E$ at time $t$, it is denoted by $\Omega_C \not \rightarrow ^t_K \Omega_E$ where $K^t\boldsymbol{\psi_\theta} \in \mathcal{F}_{E,R}$. In summary, dynamical causality and Koopman causality at time $t$ turn out to be equivalent. Namely, $\Omega_C \rightarrow ^t\Omega_E$ if and only if $\Omega_C \rightarrow^t_K \Omega_E$. The proof of this can be found in Theorem III.2 in \citet{rupeCausalDiscoveryNonlinear2024}. 

\textbf{Marginal and joint models.} The data-driven measure of Koopman causality leverages the DMD algorithm. Our method involves fitting two distinct DMD models to evolve functions in $\mathcal{F}_E$: i) a \textit{marginal model} restricted to functions from the subspace $\mathcal{F}_E$; and ii) a \textit{joint model} which utilizes functions from the joint subspace $\mathcal{F}_{E,C}$ incorporating both the effect and cause components. Our algorithm therefore requires a dataset consisting of triplets $\left( \omega_E(n), \omega_E^t(n),\omega_C(n) \right)$, where ${\omega_E^t(n):=P_E\Phi^t(\left[ \omega_E \text{ } \omega_C \text{ } \omega_R  \right]^\intercal)}$ and $n$ are the time-shift of $\omega_E$ and the data time index respectively. We split this dataset into sets of \texttt{training} and \texttt{test} triplets. 

We can now define the marginal and joint models, where the addition of identity observables on $\Omega_E$~\cite{bruntonKoopmanInvariantSubspaces2016}, $\Psi_{\text{ID},E} := {\Psi_{\text{ID}}(\Omega_E) = \Omega_E}$, empirically improves the performance of our algorithm:
\begin{equation}
    \begin{aligned}
        &K^t_{\text{marg}} = \Psi_{\text{ID},E}^t \begin{bmatrix}
            \Psi_{\text{ID},E} \\
            \Psi_{\theta,E}
        \end{bmatrix} ^\dagger ; \hspace{0.2cm}
        K^t_{\text{joint}} = \Psi_{\text{ID},E}^t \begin{bmatrix}
            \Psi_{\text{ID},E} \\
            \Psi_{\theta,E,C}
        \end{bmatrix} ^\dagger 
    \end{aligned}
\end{equation}
where the matrices $\Psi_{\theta,E}$ and $\Psi_{\theta,E,C}$ are formed from applying the marginal $\boldsymbol{\psi}_{\theta,E}(\omega_E)$ and joint $\boldsymbol{\psi}_{\theta,E,C}(\omega_{E,C})$ dictionaries 
to the \texttt{training} data  $\Omega_E$ or $\Omega_E \times \Omega_C$, respectively. 

Given this setup, our Koopman operators $K^t_{\text{marg}}$ and $K^t_{\text{joint}}$ are not square matrices. We approximate the time-shifted test function $[K^t\boldsymbol
{\psi}_{\theta,E}]([\omega_E \text{ } \omega_C \text{ } \omega_R]^T) := \omega_E^t$ through the marginal and joint models given by:
\begin{equation}
    \Tilde{w}^t_E|_{\text{marg}} :=  K^t_{\text{marg}} \begin{bmatrix}
           \boldsymbol{\psi}_{\text{ID},E}(\omega_E) \\
           \boldsymbol{\psi}_{\theta,E}(\omega_E)
         \end{bmatrix}
\end{equation}
and 
\begin{equation}
    \Tilde{w}^t_E|_{\text{joint}} :=  K^t_{\text{joint}} \begin{bmatrix}
           \boldsymbol{\psi}_{\text{ID},E}(\omega_E) \\
           \boldsymbol{\psi}_{\theta,E,C}(\omega_E,\omega_C)
         \end{bmatrix} 
\end{equation}
where $\boldsymbol{\psi}_{\text{ID},E} \in \Psi_{\text{ID},E}$.

\textbf{Causal measure.} The \textit{causal measure} $\Delta^{K^t}_{\text{C,E}}$ is defined as,
\begin{equation}\label{eq:causal_measure}
    \begin{aligned}
        \Delta^{K^t}_{\text{C,E}} &:= \Omega_C \xrightarrow[K^t]{} \Omega_E \\
        &:= \mathcal{L}(\Tilde{\omega}^t_E |_\text{marg},\omega^t_E) - \mathcal{L}(\Tilde{\omega}^t_E|_\text{joint},\omega^t_E),
     \end{aligned}
\end{equation}
where the difference in errors between the marginal and joint models is calculated over all data points $\{\omega_E, \omega_E^t, \omega_C\}$:
\begin{equation}
    \mathcal{L}(\Tilde{\omega}^t_E,\omega^t_E) = \frac{1}{N} \sum_N \left\Vert \Tilde{\omega_E^t} - \omega_E^t\right\Vert^2 .
    \label{eq:causal_measure_loss}
\end{equation}
Hence, if the joint model has a significantly lower error, it suggests that $\Omega_C$ contains causal information about $\Omega_E$. This also implies that if $\Delta^{K^t}_{\text{C,E}} \rightarrow 0$, there is minimal causal influence from $\Omega_C$ on $\Omega_E$ at time \textit{t}. 

\textbf{Conditional forecasting.} Computationally, estimating $\Delta^{K^t}_{\text{C,E}}$ at each time shift \textit{t} is expensive. Thus, we perform conditional forecasting, which preserves the structure of marginal and joint models for reasonable \textit{t}. Feeding the predicted outputs $\Tilde{\omega}_E(t)$ back as identity observables $\boldsymbol{\psi}_{\text{ID},E}(\Tilde{\omega}_E(t))$ enables efficient exploration of causal dynamics over time. However, it is not true forecasting because we supply data from the \texttt{test} dataset to construct the non-identity observables. Specifically, starting with an initial value of $\omega_E(t_0)$ from the {\color{brown} \texttt{test} data} we can define the first and following {\color{teal} conditional predictions} for the marginal model as:
\begin{equation}
        \underbrace{K^t_{\text{marg}} \begin{bmatrix}
            \boldsymbol{\psi}_{\text{ID},E}({\color{brown} \omega_E(t_0)}) \\
            \boldsymbol{\psi}_{\theta,E}({\color{brown}\omega_E(t_0)})
        \end{bmatrix}}_{:= {\color{teal} \Tilde{w}^t_E|_{\text{marg}}(t_1)}}
        \rightarrow
        \underbrace{K^t_{\text{marg}} 
        \begin{bmatrix}
            \boldsymbol{\psi}_{\text{ID},E}({\color{teal} \Tilde{\omega}_E(t_1)}) \\
            \boldsymbol{\psi}_{\theta,E}({\color{brown}\omega_E(t_1)})
        \end{bmatrix}}_{:= {\color{teal} \Tilde{w}^t_E|_{\text{marg}}(t_2)}}
        \rightarrow \dots
        \label{eq:conditional_forecasting}
\end{equation}
Given an additional initial value for $\omega_C(t_0)$ we follow the same procedure with necessary adjustments to the dictionary functions to obtain conditional forecasts for the joint model. 

\section{Experiments}
We use this section to empirically demonstrate \texttt{Kausal}'s performance and its application to identify and characterize causal mechanisms in high-dimensional nonlinear systems. Our algorithm runs with a \texttt{PyTorch} backend, and the code is available at \url{https://github.com/juannat7/kausal}. Code snippets can be found in Appendix \ref{si-sec:kausal_api} and additional details on the experimental setup and results in Appendix \ref{si-sec:experiments}, \ref{si-sec:ablation}.

\subsection{Coupled R\"{o}ssler oscillators}
We begin our exposition with the coupled R\"{o}ssler oscillators, a 6-dimensional system $\Omega \in \mathbb{R}^6$  described as:
\begin{subequations} \label{eq:coupled_rossler}
    \begin{equation}
        \begin{aligned}
            \dot{x}_1 &= -\varphi_1 y_1 - z_1, \\
            \dot{y}_1 &= \varphi_1 x_1 + a y_1 + c_1 (y_2 - y_1), \\
            \dot{z}_1 &= b + z_1 (x_1 - d).
        \end{aligned}
    \end{equation}
    \begin{equation}
        \begin{aligned}
            \dot{x}_2 &= -\varphi_2 y_2 - z_2, \\
            \dot{y}_2 &= \varphi_2 x_2 + a y_2 + c_2 (y_1 - y_2), \\
            \dot{z}_2 &= b + z_2 (x_2 - d)
        \end{aligned}
    \end{equation}
\end{subequations}
where $a, b, d, \varphi_i \in \mathbb{R}$ are prescribed parameters and $c_i \in \mathbb{R}$ the coupling terms. Setting $c_2 = 0$ naturally partitions the variables into $\Omega_C = [x_2, y_2, z_2]^\intercal \rightarrow^t_K \Omega_E = [x_1, y_1, z_1]^\intercal$. Here, we use multilayer perceptrons (MLP) as our deep estimators for the observables, represented as a $M=32$-dimensional feature vector of the encoder final layer's output (see Appendix \ref{si-sec:experiments_coupled_rossler}).

\begin{figure}[h!]
    \centering
    \includegraphics[width=\linewidth]{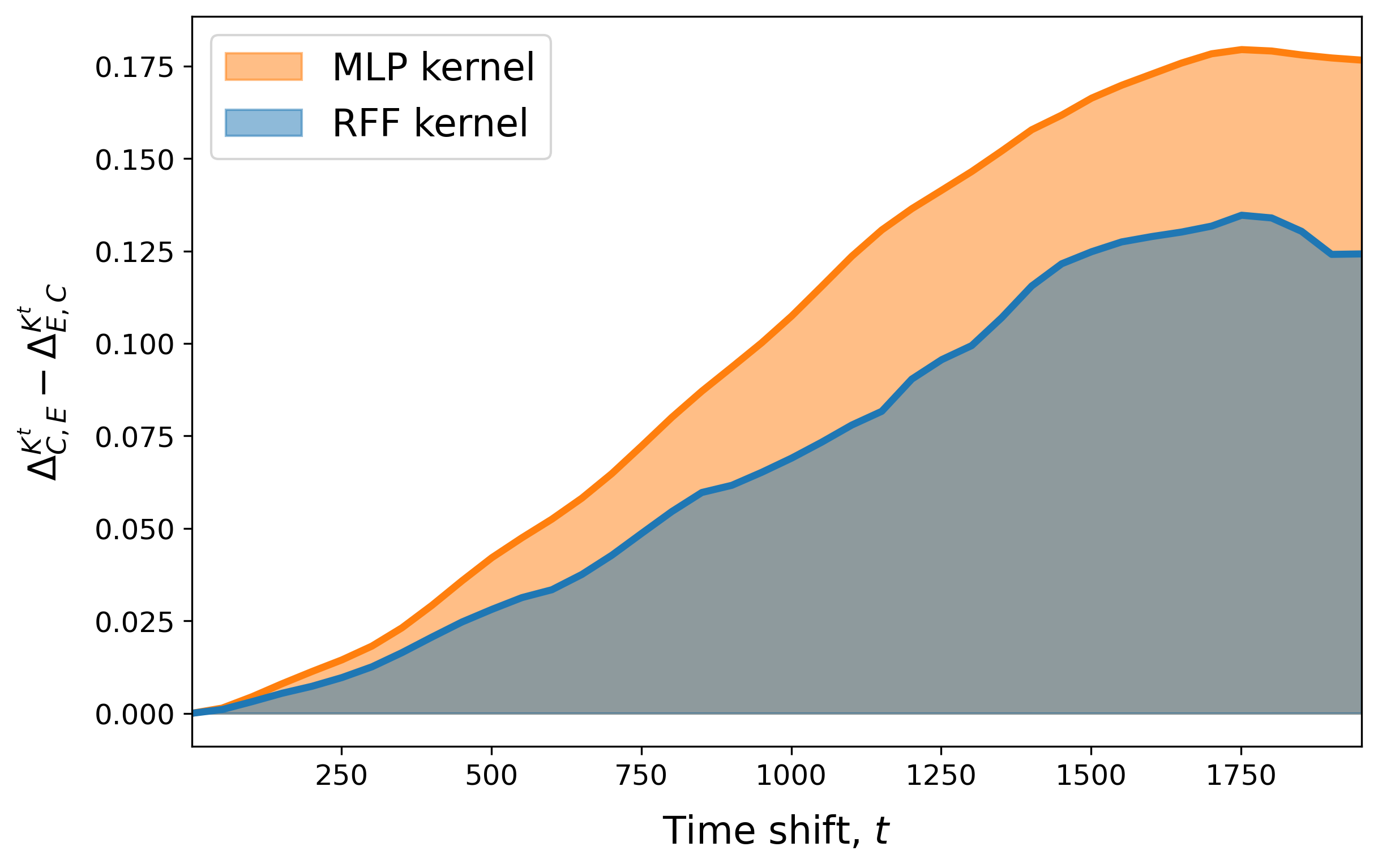}
    \caption{Causal measure estimation of coupled R\"{o}ssler oscillators by computing the difference between the true causal ($\Delta^{K^t}_{C,E}$) and non-causal direction ($\Delta^{K^t}_{E,C}$) across time shifts $t$, using different kernels to approximate the observables.}
    \label{fig:coupled_rossler_estimation}
\end{figure}

\textbf{Causal measure estimation.} First, we estimate causal measures as defined in Equation \ref{eq:causal_measure_loss} where the measure for each unique time shift $t \in \mathbb{Z}_{+}$ is estimated across $N$ data points. As illustrated in Figure \ref{fig:coupled_rossler_estimation}, \texttt{Kausal} is able to evaluate statistically significant\footnote{See Algorithm~\ref{alg:hypo_test} with $\rho_{crit} = 0.05$.} causal signals in the true direction ($\Delta^{K^t}_{C,E}$) when compared with the non-causal case ($\Delta^{K^t}_{E,C}$). For completeness, we also plot the causal measure using an RFF kernel ($M=500$) following~\citet{rupeCausalDiscoveryNonlinear2024}.

\begin{figure}[h!]
    \centering
    \begin{subfigure}[b]{0.5\textwidth}
        \centering
        \includegraphics[width=\textwidth]{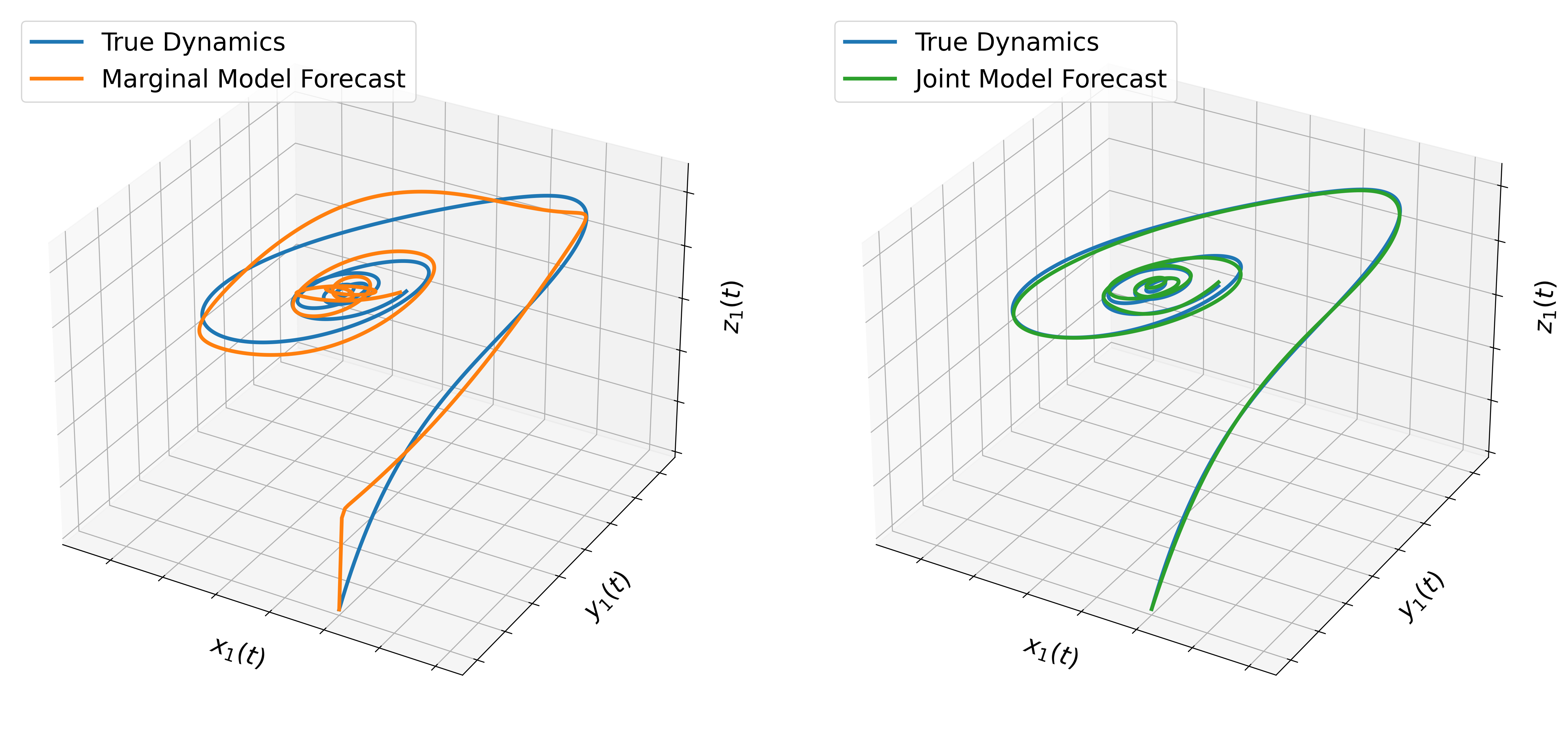}
        \caption{Conditional forecasts in the \textit{true} causal direction $\Omega_{C} \rightarrow^t_K \Omega_E$}
    \end{subfigure}
    \hfill
    \begin{subfigure}[b]{0.5\textwidth}
        \centering
        \includegraphics[width=\textwidth]{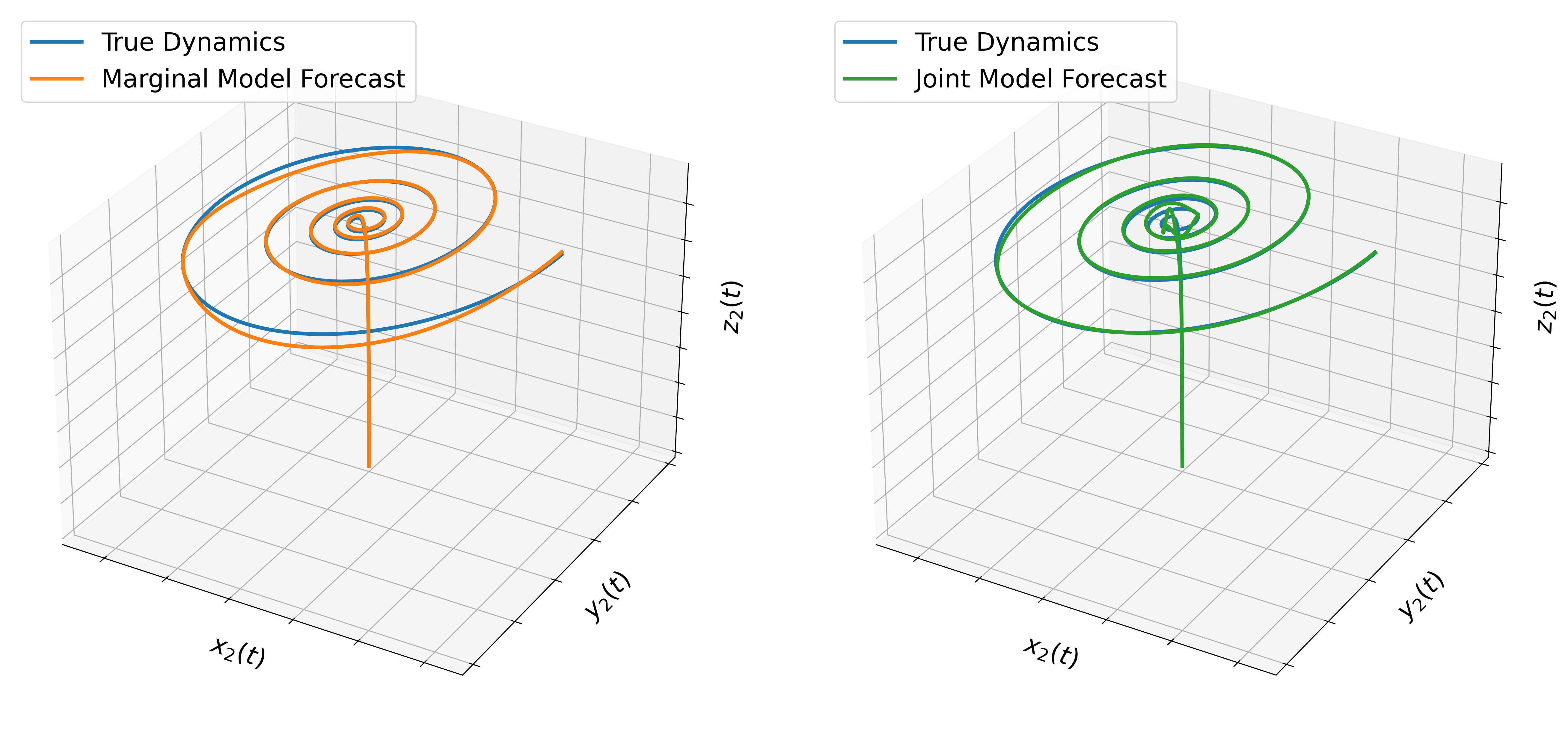}
        \caption{Conditional forecasts in the non-causal direction $\Omega_{E} \rightarrow^t_K \Omega_C$}
    \end{subfigure}
    \caption{Conditional forecasts in the (a) \textit{true} and (b) non-causal direction using MLP kernels. In (a), the addition of $\Omega_C$ in the joint model improves the forecast of $\Omega_E$ relative to the marginal model that excludes it. In (b), however, both marginal and joint models make no qualitative difference as $\Omega_E \not \rightarrow^t_K \Omega_C$.}
    \label{fig:coupled_rossler_forecasting}
\end{figure}

\textbf{Performing conditional forecasting.} We also illustrate the concept of conditional forecasting formalized earlier in Equation \ref{eq:conditional_forecasting}, starting with $\omega_0$. As shown in Figure \ref{fig:coupled_rossler_forecasting}, the inclusion of $\Omega_C$ (i.e., the joint model) is crucial in capturing $\Omega_E$ dynamics in the \textit{true} causal direction, whereas its exclusion (i.e., the marginal model) leads to significant deviations. Conversely, in the non-causal direction, both the marginal and joint models make no perceptible difference in forecasting $\Omega_C$ as $\Omega_E \not \rightarrow^t_K \Omega_C$.

\begin{figure}[h!]
    \centering
    \begin{subfigure}[b]{0.5\textwidth}
    \includegraphics[width=\linewidth]{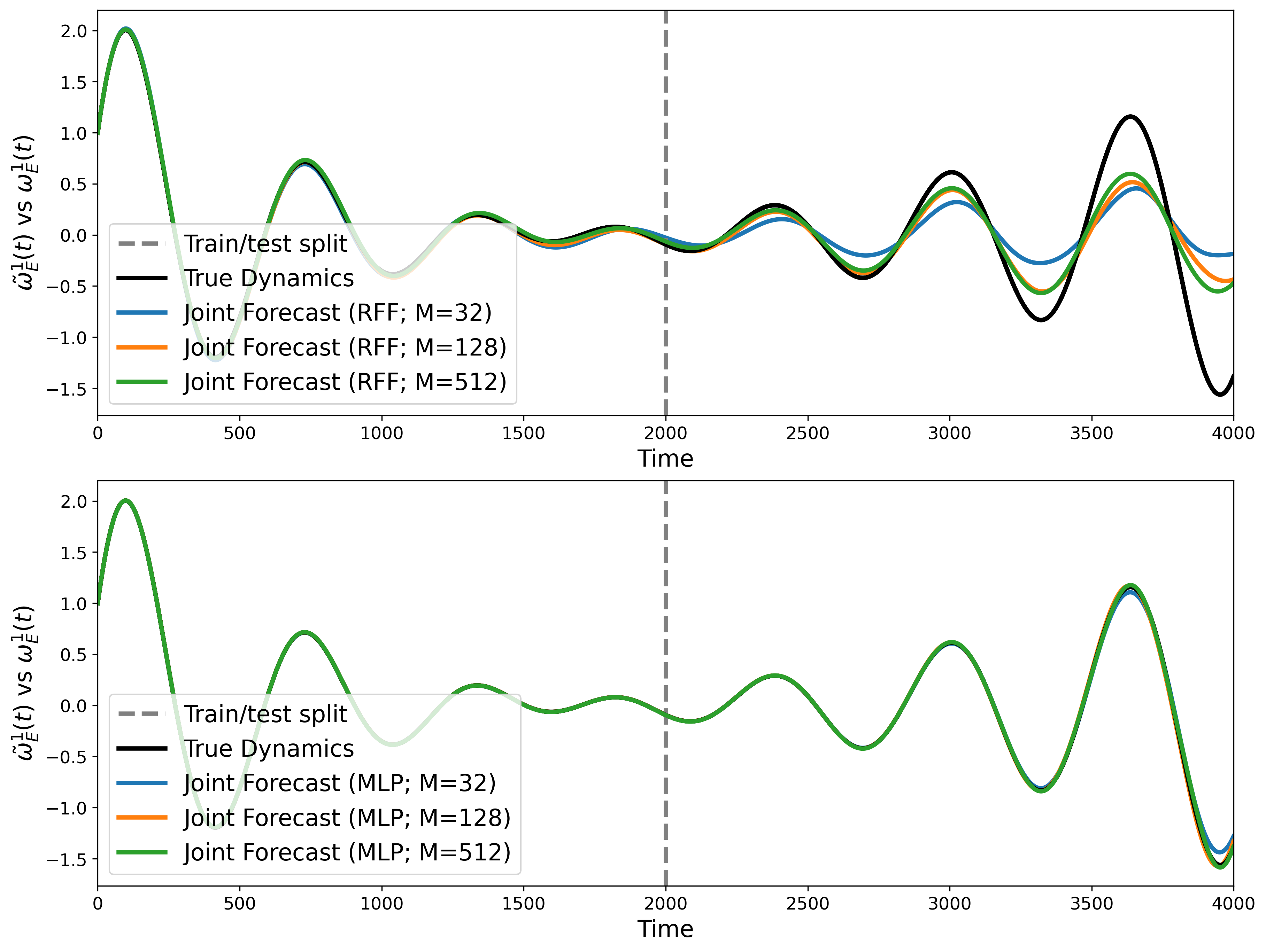}
    \caption{Ablating optimal observables by performing conditional joint forecasts (for ``fixed" $K^{t=1}$) in the $\Omega_C \rightarrow^t_K \Omega_E$ direction (${\omega^1(t) := x_1(t) \in \Omega_E}$) using different estimators of varying $M$.}
    \end{subfigure}
    \hfill
    \begin{subfigure}[b]{0.5\textwidth}
    \centering
    \includegraphics[width=\linewidth]{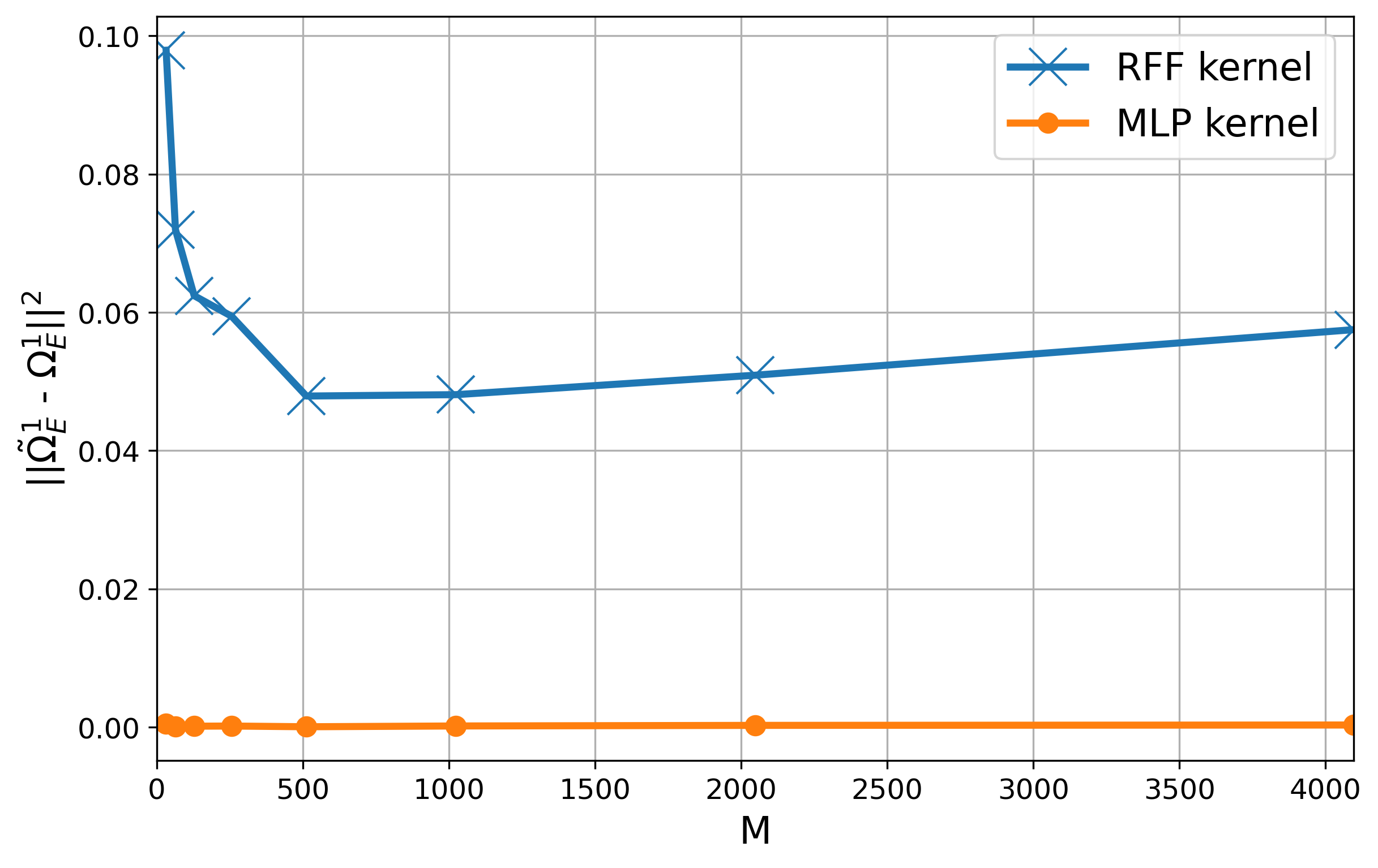}
    \caption{Dimensionality scaling between prescribed $\boldsymbol{\psi}_\text{RFF}$ and learnable $\boldsymbol{\psi}_\text{MLP}$ comparing the conditional forecasts' MSE across all times.}
    \end{subfigure}
    \caption{Performance assessment of MLP and RFF kernels for varying dimensionality (i.e. increasing complexity).}
    \label{fig:coupled_rossler_ablate_M}
\end{figure}

\textbf{Optimal kernel selection.} Robust estimation of observables is crucial for the accurate representation of nonlinear dynamics in $\mathcal{F}$. In order to study this notion, we perform joint forecasts using either a prescribed $\boldsymbol{\psi}_\text{RFF}$ or a learnable dictionary of functions, $\boldsymbol{\psi}_\theta$, for varying dimensionality $M$. As illustrated in Figure~\ref{fig:coupled_rossler_ablate_M}, joint forecasts utilizing $\boldsymbol{\psi}_\theta$ reproduce the true dynamics more accurately (expected behavior), even beyond the \texttt{training} set and for low $M$. In contrast, joint forecasts using $\boldsymbol{\psi}_\text{RFF}$, even with high $M$, struggle to capture the underlying dynamics, underscoring the limitations of prescribing dictionary functions rather than directly learning them from data.

\subsection{Reaction-diffusion equation}
We now showcase \texttt{Kausal}'s scalability to high-dimensional settings and underscore the notion of uncertainty quantification of causal measures using a 2D nonlinearly coupled reaction-diffusion process described as:
\begin{equation}
\label{eq:reaction_diffusion}
\begin{aligned}
    \dot{u} &= D_u \nabla^2 u - u (u - a)(u - 1) + \beta v , \\
    \dot{v} &= D_v \nabla^2 v - v (v - b)(v - 1) + \gamma u 
\end{aligned}
\end{equation}
where $D_u, D_v, a, b \in \mathbb{R}$ are prescribed parameters, ${\nabla^2: \mathbb{R}^{n_x, n_y} \rightarrow \mathbb{R}^{n_x, n_y}}$ the 2D Laplace operator (${n_x = n_y = 16}$), and $\beta, \gamma \in \mathbb{R}$ the coupling terms. The states ($u,v$) include components of the horizontal velocity. Here, we set $\gamma = 0$, such that the variables are naturally partitioned into $\Omega_C = [v]^\intercal \rightarrow^t_K \Omega_E = [u]^\intercal$. In our setup, we use convolutional neural networks (CNN) as our deep estimators for the observables, represented as a $128$-dimensional feature map of the encoder final layer's output (see Appendix~\ref{si-sec:experiments_reaction_diffusion} for additional details).

\begin{figure}[h!]
    \centering
    \includegraphics[width=\linewidth]{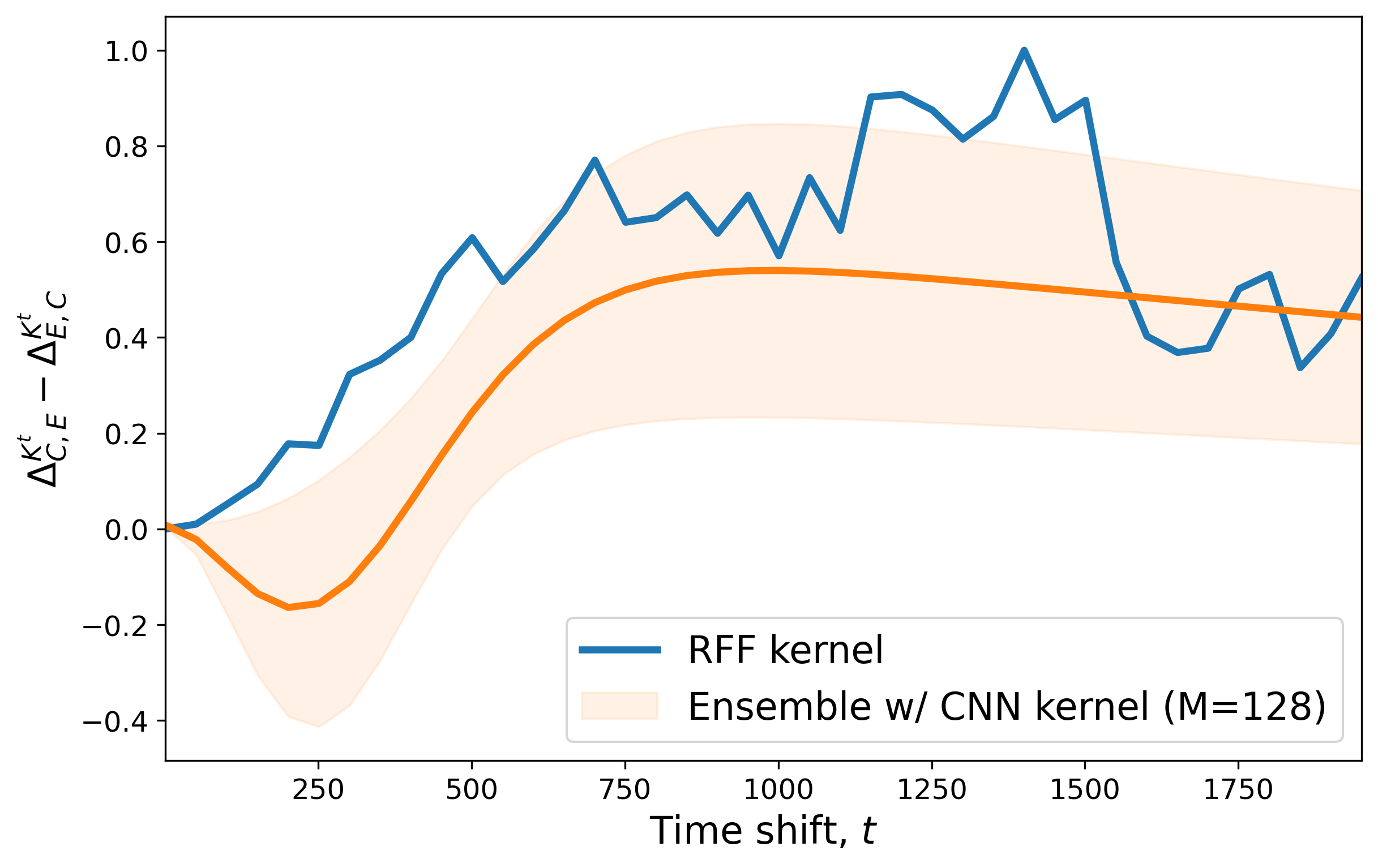}
    \caption{Causal measure of reaction-diffusion equation by computing the difference between the true causal ($\Delta^{K^t}_{C,E}$) and non-causal direction ($\Delta^{K^t}_{E,C}$) across time shifts $t$, using an ensemble of randomly initialized CNN kernels ($M=128$) to estimate the observables, versus the default RFF.}
    \label{fig:reaction_diffusion_estimation}
\end{figure}

\textbf{Uncertainty quantification of causal measures}. Our finite approximation of linear flow in $\mathcal{F}$ is imperfect, and therefore, capturing its representation uncertainty is crucial. To this end, we fit a number of CNN kernels to form an ensemble, where we maintain the dimensionality ($M = 128$) but randomly initialize the learnable weights. In Figure~\ref{fig:reaction_diffusion_estimation}, \texttt{Kausal} is able to extract statistically significant causal signals in the true direction ($\Omega_{C} \rightarrow^t_K \Omega_E$) when compared with the non-causal case ($\Omega_{E} \rightarrow^t_K \Omega_C$) as $t \gg 0$. The ensemble spread is represented as the shaded region around the mean solid line. Note that due to small state values, we scale the causal measure to $[-1,1]$ and negative values of the causal measure are due to an insignificant causal signal.

\subsection{ENSO model} 
As a final experiment, we illustrate \texttt{Kausal}'s generalization to real-world phenomena using an example from climate science, El Ni\~no–Southern Oscillation (ENSO). ENSO is an important source of predictability in the climate system \cite{wangUnderstandingEnsoPhysicsA2013}. As such, a robust understanding of its mechanism and how it evolves in a changing climate is of widespread importance. Due to its major societal impacts, this is especially true for explainable decision making in critical sectors such as agriculture and disaster preparedness \citep[e.g.,][]{glantzSocietalImpactsAssociated1987,callahanPersistentEffectNino2023}. 

We first apply \texttt{Kausal} to analyze a well-known ENSO physics-based dynamical model \cite{jinEquatorialOceanRecharge1997} defined as:
\begin{equation}
\label{eq:enso}
\begin{aligned}
    \dot{T} &= -r T - \mu \alpha b_0 h - \epsilon T^3 , \\
    \dot{h} &= \gamma T + (\gamma \mu b_0 - c) h
\end{aligned}
\end{equation}
where $r, \alpha, b_0, c \in \mathbb{R}$ are prescribed parameters, $\epsilon \in \mathbb{R}$ is a damping coefficient regulating the strength of nonlinearity, and $\gamma, \mu \in \mathbb{R}$ are the coupling terms. The states include anomalies in sea surface temperature $T$ and thermocline height $h$ in the eastern and western equatorial Pacific, respectively. For notational uniformity, we partition $\Omega_C = [T]^\intercal$ and $\Omega_E = [h]^\intercal$. However, this does not imply $\Omega_E \not \rightarrow^t_K \Omega_C$ as long as the coupling terms are non-zero, which is the case throughout our exposition. In our setup, we use MLP as our deep estimator for the observables, represented as a $32$-dimensional vector of the encoder final layer's output.

\begin{figure}[!h]
    \centering
    \begin{subfigure}[b]{0.5\textwidth}
        \centering
        \includegraphics[width=\textwidth]{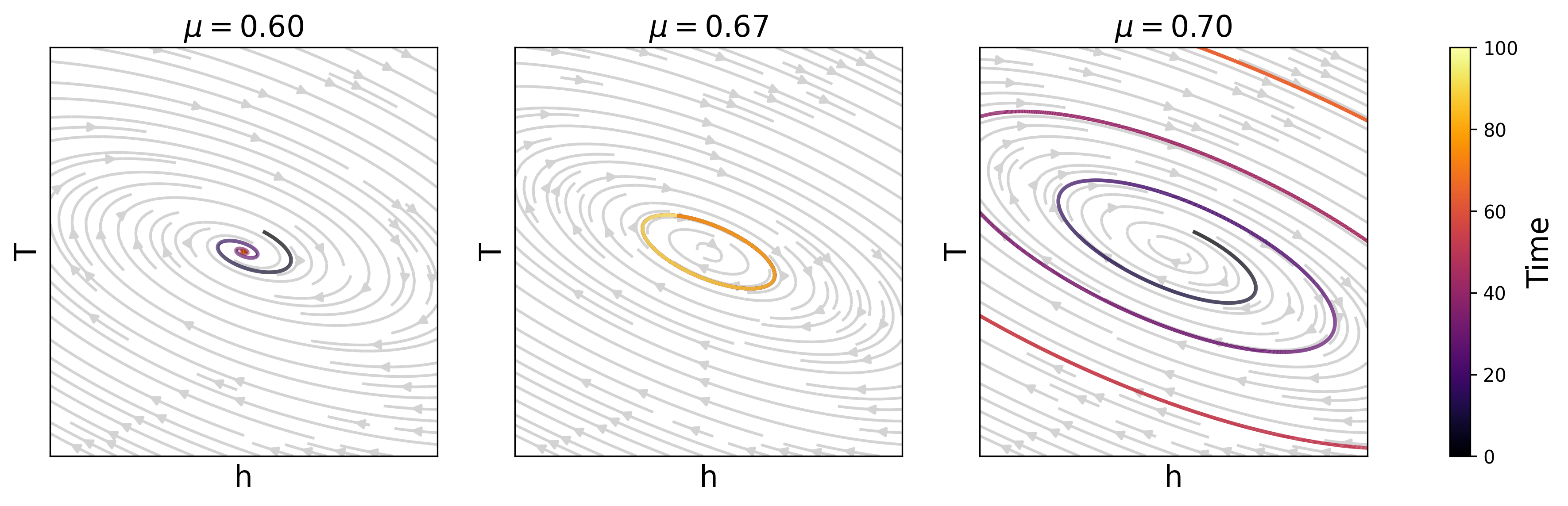}
        \caption{Phase space of ENSO dynamics in different $\mu$-regimes.}
    \end{subfigure}
    \hfill
    \begin{subfigure}[b]{0.5\textwidth}
        \centering
        \includegraphics[width=\textwidth]{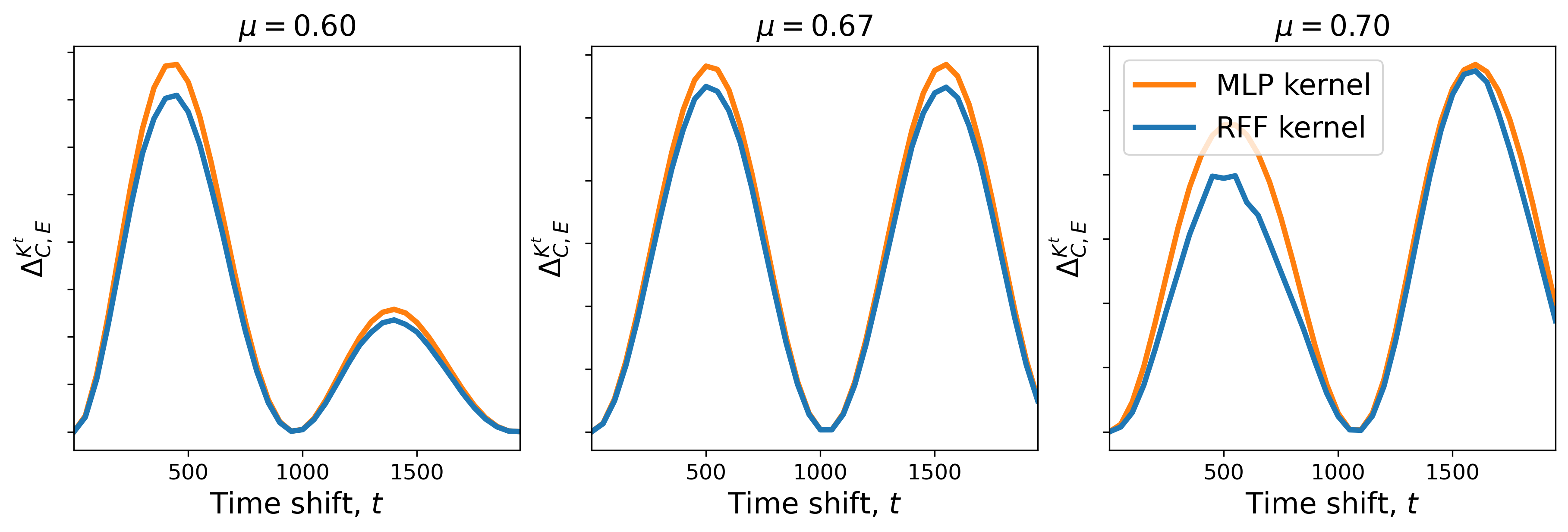}
        \caption{Causal measure estimation of ENSO ($\Delta^{K^t}_{T,h}$) across $\mu$-regimes using different estimators for the observables.}
    \end{subfigure}
    \caption{\texttt{Kausal} reveals the underlying ENSO dynamics exhibiting (\textit{left}) dissipative, (\textit{center}) stable, and (\textit{right}) diverging oscillatory patterns shown in (a) the phase space, and (b) causal measure.}
    \label{fig:enso_estimation}
\end{figure}

\textbf{Linear stability analysis}. Setting $\epsilon = 0$ in Equation~\ref{eq:enso} linearizes the dynamics such that several evaluations could be performed, including stability analysis. This is crucial for identifying bifurcation, or tipping points of the system. The full linear stability analysis, including the bifurcation diagram, are described and illustrated in Figure~\ref{si-fig:enso_linear_analysis}. Given a prescribed set of parameters, the critical point ${\mu_c = 0.67}$ corresponds to a stable periodic attractor.
Meanwhile, $\mu < \mu_c$ and $\mu > \mu_c$ represent dissipative and diverging oscillations, respectively. Here, we only consider physically realistic oscillatory dynamics defined in $\mu$-regimes where $\text{Im}(\lambda) \neq 0$.

In Figure~\ref{fig:enso_estimation}, \texttt{Kausal} reveals accurate linear analysis across $\mu$-regimes that induces dissipative, stable, or diverging oscillatory ENSO dynamics. For instance, in the critical delta region where $\mu_c = 0.67$, our causal measures show identical peaks with consistent periodicity. Whereas in the dissipative (or diverging) dynamics, our causal measures reveal decreasing (or increasing) influence in the $\Omega_C \rightarrow^t_K \Omega_E$ direction. This is physically consistent with the model as both $T$ and $h$ are exerting coupled feedback, and partly governed by $\mu$. 

\begin{figure}[h]
    \centering
    \includegraphics[width=\linewidth]{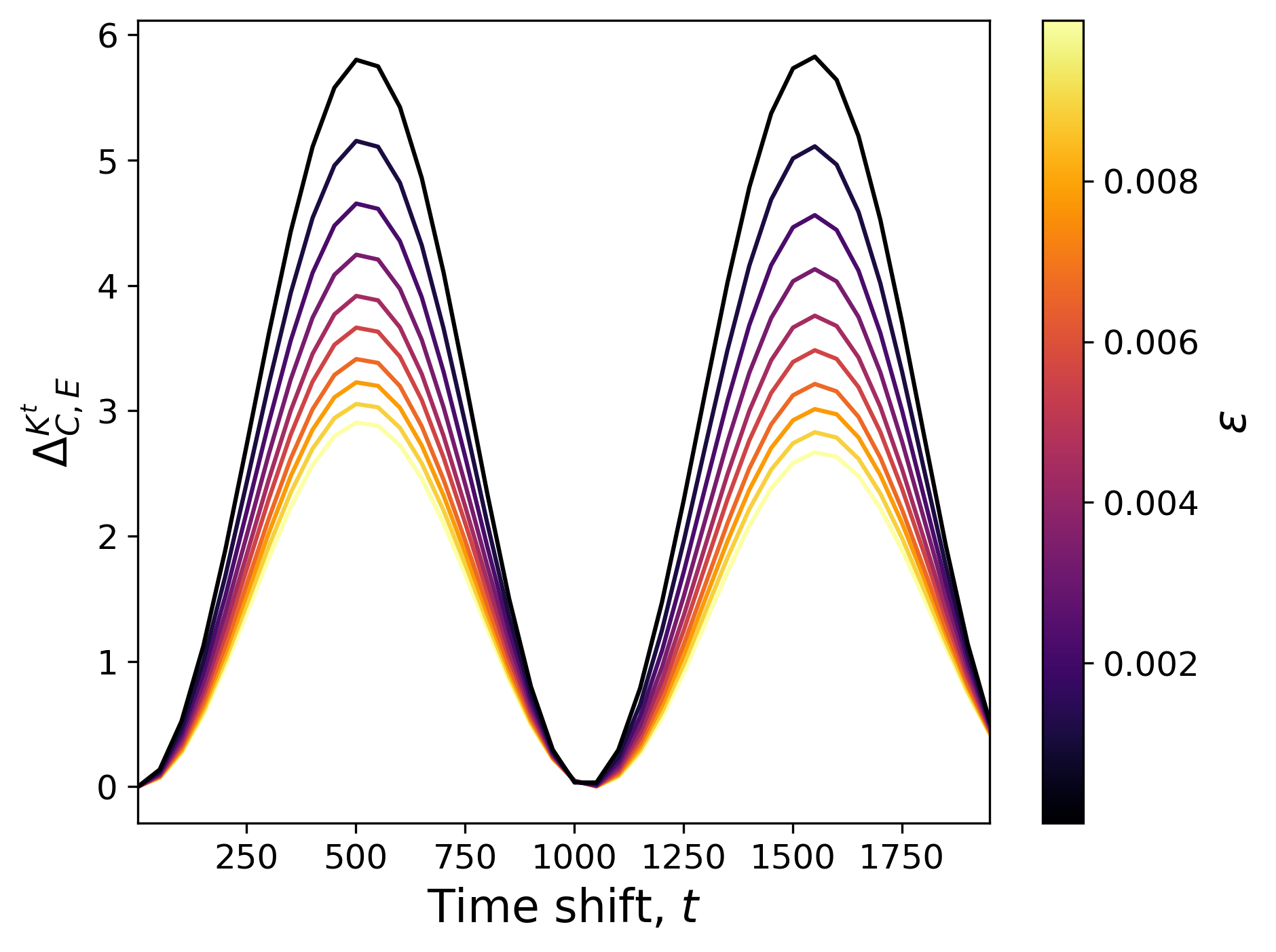}
    \caption{Causal measure of ENSO for varying nonlinearity strength $\epsilon$ using MLP kernels to approximate the observables. Our \texttt{Kausal} framework is able to meaningfully differentiate and characterize nonlinearity in the underlying ENSO dynamics.}
    \label{fig:enso_estimation_mlp}
\end{figure}

\textbf{Capturing nonlinearity strength}. We perform similar analysis in the stable region of $\mu_c = 0.67$, but vary nonlinearity by setting $\epsilon \neq 0$. As illustrated in Figure~\ref{fig:enso_estimation_mlp}, we note two physically consistent observations in the causal measure: (a) \textit{oscillatory dynamics} (insofar $\text{Im}(\lambda) \neq 0$), and (b) a \textit{dampening pattern} as nonlinearity strengthens, as evidenced by the slower rate of change in $\Delta^{K^t}_{C,E}$ and smaller magnitude of peaks. Overall, a robust identification of weakly or strongly coupled variables within a system, as demonstrated by \texttt{Kausal}, is crucial for better understanding, effective control, and accurate forecasting of dynamics.

\begin{figure*}[!h]
    \centering
    \includegraphics[width=0.9\linewidth]{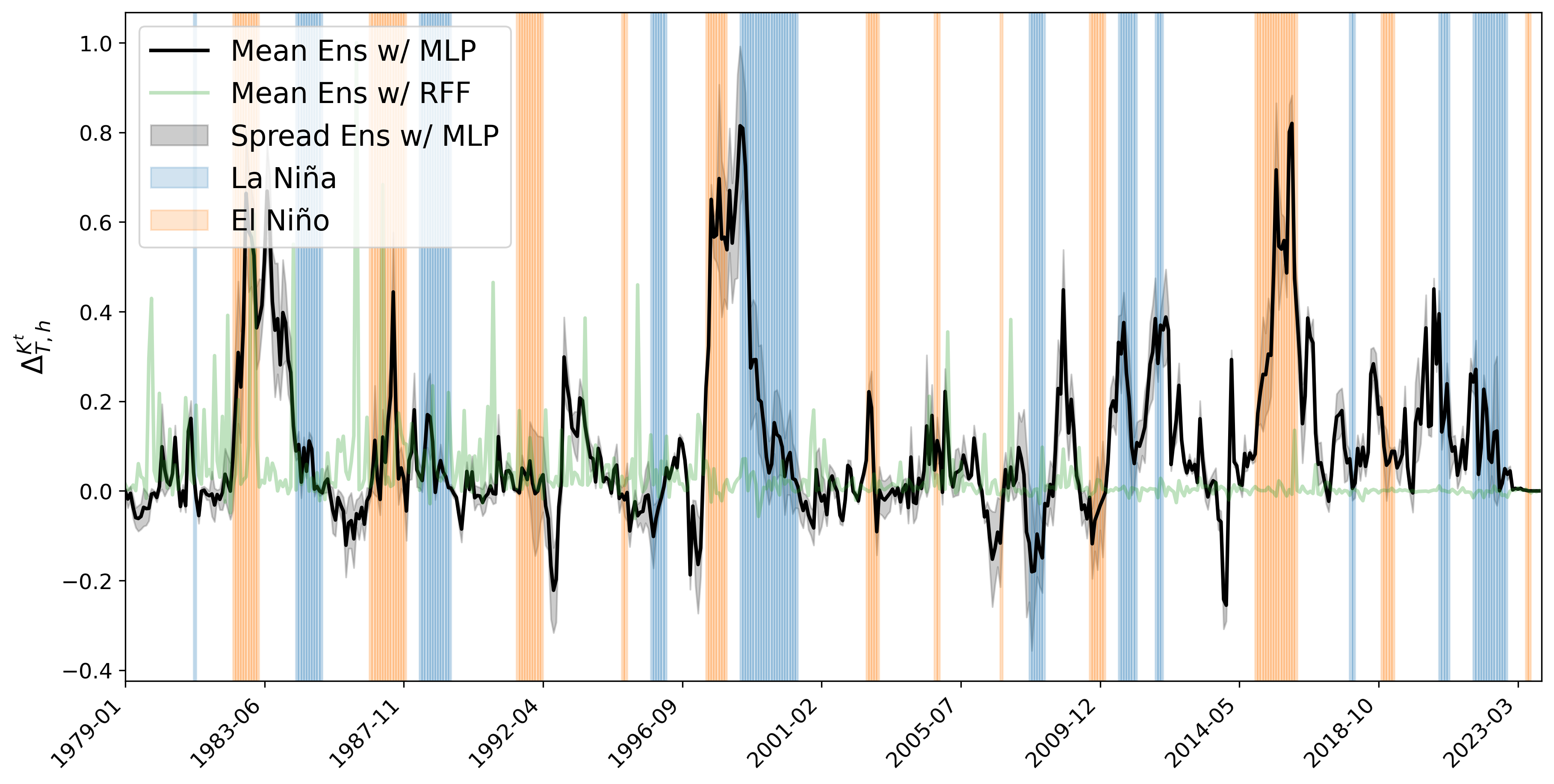}
    \caption{Causal measure of ENSO using real-world data showcasing \texttt{Kausal}'s ability (black line) to capture major El Ni\~no (red shading) and La Ni\~na (blue shading) events. Events are estimated using NOAA's Oceanic Ni\~no Index \cite{glantz2020OniIndex}. The causal measure estimated using a prescribed RFF kernel (green line) fails to capture ENSO dynamics.}
    \label{fig:enso_real_mlp}
\end{figure*}

\subsection{\texttt{Kausal} in the wild}
Lastly, we demonstrate the feasibility of \texttt{Kausal} to study ENSO dynamics using ocean reanalysis products from the European Centre for Medium-Range Weather Forecasts \citep[ECMWF;][]{zuo2019ecmwf}. Real-world dataset are often sparse and noisy \cite{nathaniel2023metaflux,kim2024spatiotemporal}, and leveraging physics-informed priors through causal discovery has been shown to be useful for various tasks including for emulation \cite{iglesias2024causally}. Preliminary results as illustrated in Figure~\ref{fig:enso_real_mlp} are promising. We find that \texttt{Kausal} is able to track the emergence of ENSO events as represented by $\Delta^{K^t}_{T,h} > 0$. Specifically, warm-phase El Ni\~no events (red shading in Figure~\ref{fig:enso_real_mlp}) result from an increase in $T$, weakening the Walker circulation and leading to a decrease in $h$. Cold-phase La Ni\~na events follow a similar pattern but in the opposite direction (blue shading in Figure~\ref{fig:enso_real_mlp}). See Appendix~\ref{si-sec:experiments_enso} for more details. The estimated causal measure captures major El Ni\~no events in 1982--83, 1997--98, and 2014--16, representing some of the strongest in recorded history~\cite{glantzSocietalImpactsAssociated1987, mcphadenGenesisEvolution1997981999, huExtremeNino201520162017}.
Additionally, Figure~\ref{fig:enso_real_mlp} highlights \texttt{Kausal}'s improved performance with respect to existing work that uses prescribed dictionary functions~\citep[e.g.,][]{wang2022koopman,rupeCausalDiscoveryNonlinear2024}.

This analysis represents a first step in contributing to ongoing research on possible state changes of ENSO due to increasing ocean temperatures which could have major global impacts~\citep[e.g.,][]{aiken1970sShiftENSO2013, bayrNinoSouthernOscillationTipping2024}.

\subsection{Baselines}

In the following, we compare \texttt{Kausal} to a range of baseline causal discovery methods across two tasks: (i) causal direction identification, and (ii) causal magnitude estimation. Specifically, we compare against two well-established methods: PCMCI+ \citep[Peter Clark Momentary Conditional Independence;][]{runge2020discovering} and VARLiGAM \cite{hyvarinen2010estimation}, which combines the basic LiNGAM model \citep[Linear Non-Gaussian Acyclic Model;][]{shimizu2006linear} with vector autoregressive models (VAR) for time series applications. Further, we evaluate against other novel deep learning-based causal discovery methods: TSCI \citep[Tangent Space Causal Inference;][]{butler2024tangent}, where causality is detected if a map exists between the reconstructed state spaces representing different dynamical systems; cLSTM \citep[Neural Granger Causality;][]{tank2021neural}, which learns the causal structure using long short-term memory (cLSTM) to model nonlinear autoregressive relationships within time series. See Appendix~\ref{si-sec:ablation} for more details. 

\textbf{Task 1: Identify causal direction.} We first conduct hypothesis testing to identify the causal direction for the coupled R\"{o}ssler oscillators (Table~\ref{tab:cro}) and the reaction-diffusion equation (Table~\ref{tab:rde}) for which the true directionality is known and clearly distinguishable (more details in Appendix~\ref{app:task1}). Because none of the deep learning-based methods provide out-of-the-box statistical tests to validate causal direction, we conduct this task with VARLiNGAM and PCMCI+.

\begin{table}[h!]
    \caption{Causal direction identification for (a) Coupled R\"{o}ssler Oscillators, (b) reaction-diffusion equation. Entries denote the p-value ($\rho$) for the true (C $\rightarrow$ E; $\downarrow\rho$ is better) and false causal direction (E $\rightarrow$ C; $\uparrow\rho $ is better).}
    \label{combined-table}
    \begin{center}
    \begin{small}
    \begin{sc}
    
    \begin{subtable}{0.49\textwidth}
        \centering
        \caption{Coupled R\"{o}ssler Oscillators}
        \label{tab:cro}
        \begin{tabular}{lcc}
            \toprule
            Method & $\rho_{\text{C} \rightarrow \text{E}}$ & $\rho_{\text{E} \rightarrow \text{C}}$ \\
            \midrule
            PCMCI+  & $0.44$ & $0.39$ \\
            VARLiNGAM  & $1.22 \times 10^{-5}$ & $0.00$\\
            \texttt{Kausal}$_\text{RFF}$ & $\mathbf{0.01}$ & $\mathbf{1.00}$ \\
            \texttt{Kausal}$_\text{MLP}$ & $\mathbf{0.01}$ & $\mathbf{1.00}$ \\
            \bottomrule
        \end{tabular}
    \end{subtable}
    
    \vskip 0.1in  
    
    \begin{subtable}{0.49\textwidth}
        \centering
        \caption{Reaction-diffusion equation}
        \label{tab:rde}
        \begin{tabular}{lcc}
            \toprule
            Method & $\rho_{\text{C} \rightarrow \text{E}}$ & $\rho_{\text{E} \rightarrow \text{C}}$ \\
            \midrule
            PCMCI+  & $0.74$ & $1.95 \times 10^{-13}$ \\
            VARLiNGAM  & $0.00$ & $0.00$\\
            \texttt{Kausal}$_\text{RFF}$ & $0.03$ & $0.03$ \\
            \texttt{Kausal}$_\text{CNN}$ & $\mathbf{0.03}$ & $\mathbf{0.23}$ \\
            \bottomrule
        \end{tabular}
    \end{subtable}
    
    \end{sc}
    \end{small}
    \end{center}
    \vskip -0.1in
\end{table}

Our results indicate that PCMCI+ tends to be conservative, often yielding no directionality when there is. In contrast, VARLiNGAM frequently conflates directional cues, inferring bi-directional relationships when there is none. \texttt{Kausal}, however, captures the true (and the lack of) causal relationship with greater precision.

\begin{figure}[!t]
    \centering
    \includegraphics[width=0.8\linewidth]{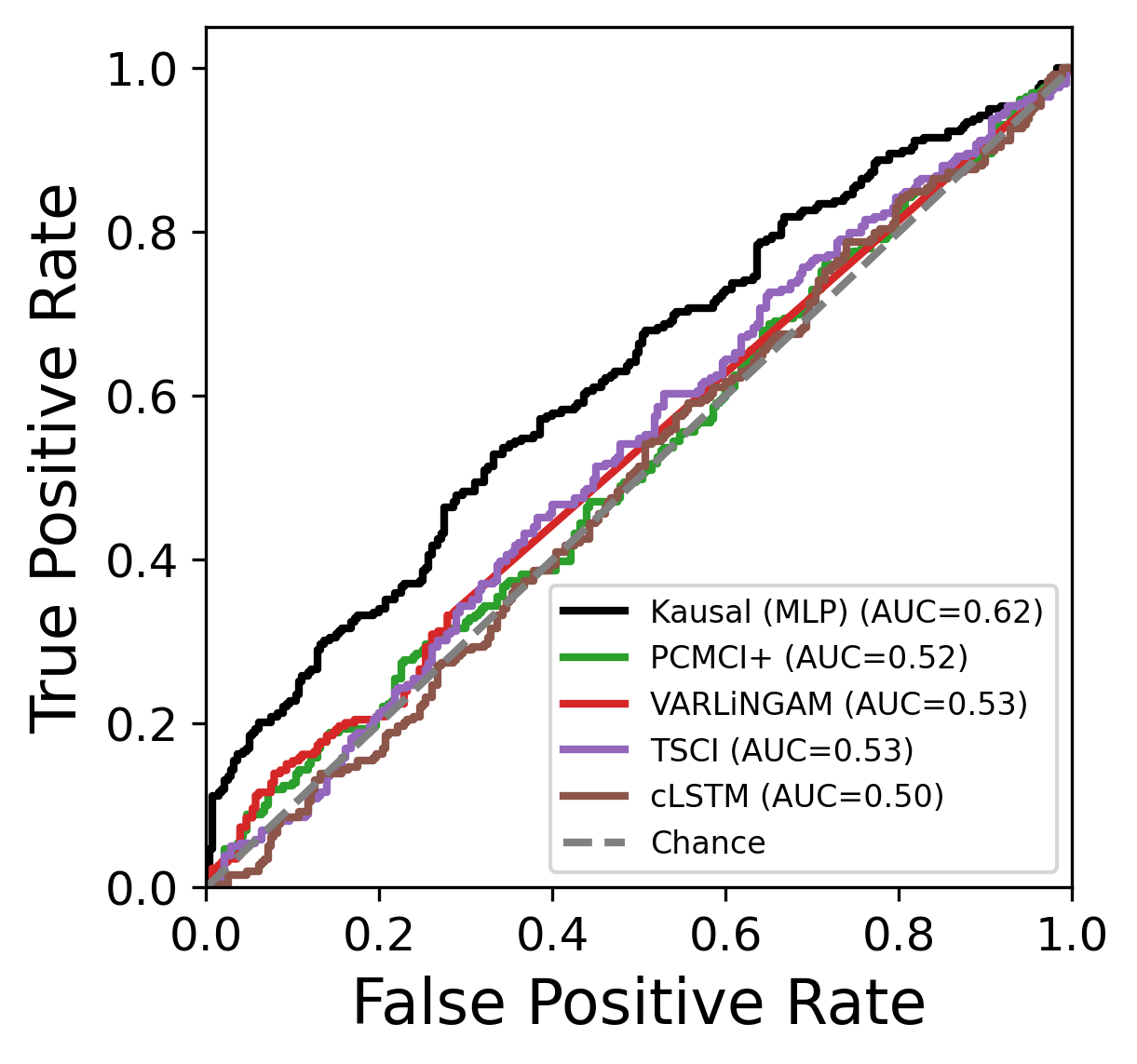}
    \caption{AUROC scores for the observed ENSO (AUROC $> 0.5$ shows significant skills better than chance).}
    \label{fig:auroc-real-enso}
\end{figure}

\textbf{Task 2: Evaluate causal magnitude.} Next, we evaluate \texttt{Kausal} for causal magnitude estimation. This analysis extends the observed ENSO case (Figure~\ref{fig:enso_real_mlp}) but added a summarized quantitative evaluation in terms of Area Under the Receiver Operating Characteristic \citep[AUROC;][]{peterson1954theory} scores (Figure~\ref{fig:auroc-real-enso}). Since most, if not all existing methods still assume a static causal graph with fixed link magnitudes, we employ the sliding‐window protocol introduced by \citet{runge2019inferring} for their ENSO analysis (see Appendix~\ref{app:task2} for details and results). Beyond the ENSO case, we further apply similar setup to predict causal magnitudes in other toy problems. For instance, Figures \ref{si-fig:coupled_rossler_estimation} - \ref{si-fig:enso_real_estimation} illustrate the causal magnitude timeseries given a backdrop of extremes to be detected. While Figures \ref{si-fig:coupled_rossler_auroc}-\ref{si-fig:enso_real_auroc} highlight the corresponding AUROC scores.

Overall, \texttt{Kausal} consistently achieves the highest score. The GC‐based approaches (PCMCI+, cLSTM and VARLiNGAM) fare worse because their fixed time‐lagged embeddings cannot capture nonlinearities and multiscale interactions outside the chosen window. Although TSCI leverages cross‐correlation maps in an operator‐theoretic approach, it still struggles due to the absence of explicit linearity constraints.

\section{Conclusions}
This paper presents \texttt{Kausal}, a novel deep Koopman operator-based algorithm for causal discovery in high-dimensional nonlinear systems, combining the principles of deep learning, Koopman theory, and causal inference. Leveraging deep learning to infer the dictionary functions of observables, our algorithm improves the estimation of causal measures in nonlinear dynamics relative to approaches that prescribe dictionary functions. \texttt{Kausal}'s scalability and generalizability is further highlighted by evaluating its performance for a range of systems of increasing complexity, including coupled R\"{o}ssler oscillators, the reaction-diffusion equation, as well as a toy model and observations of ENSO.

There are several promising directions that build on our proposed deep Koopman causal discovery framework. First, we will extend the framework to causal effect estimation allowing us to generate explainable physics-informed predictions that are robust to out-of-distribution scenarios. Second, we can extend \texttt{Kausal} to discover structural causal graph by developing tractable adjacency testing algorithms. Third, we will also study stochastic, non-autonomous dynamics accounting for intrinsic variability and time-dependent flow. 


\section*{Software and Data}
We release the code at: \url{https://github.com/juannat7/kausal}.

\section*{Acknowledgments}
The authors acknowledge funding, computing, and storage resources from the NSF Science and Technology Center (STC) Learning the Earth with Artificial Intelligence and Physics (LEAP) (Award \#2019625), and Department of Energy (DOE) Advanced Scientific Computing Research (ASCR) program (DE-SC0022255). This work was supported in part by the U.S. DOE, Office of Science, Office of Biological and Environmental Research, Regional and Global Model Analysis program area as part of the HiLAT-RASM project. The Pacific Northwest National Laboratory (PNNL) is operated for DOE by Battelle Memorial Institute under contract DE-AC05-76RLO1830. J.B., K.L., and P.G. acknowledge funding from the Zegar Family Foundation. J.N. also acknowledges funding from Columbia-Dream Sports AI Innovation Center.


\bibliographystyle{icml2025}
\bibliography{references}

\appendix
\renewcommand{\thefigure}{S\arabic{figure}}
\renewcommand{\thetable}{S\arabic{table}}
\setcounter{figure}{0} 
\setcounter{table}{0}  

\onecolumn
\onecolumn
{
    \centering
    \Large
    \textbf{Deep Koopman operator framework for causal discovery \\ in nonlinear dynamical systems}\\
    \vspace{0.5em}Supplementary Material \\
    \vspace{0.5em}
    {\normalsize
      \makebox[0.95\textwidth]{%
        \textbf{Juan Nathaniel}\textsuperscript{*}, 
        \textbf{Carla Roesch}\textsuperscript{*},%
      }\\[0.3em]
      \makebox[0.95\textwidth]{%
        \textbf{Jatan Buch}, 
        \textbf{Derek DeSantis}, \textbf{Adam Rupe}, 
        \textbf{Kara Lamb},
        \textbf{Pierre Gentine}%
      }\\[0.5em]
    }
}
\renewcommand{\thefootnote}{\fnsymbol{footnote}}
\footnotetext[1]{Corresponding authors: jn2808@columbia.edu,
cmr2293@columbia.edu}

\section{Relevant aspects of reproducing kernel Hilbert spaces}\label{sec:math}

\begin{table}[h!]
\centering
\caption{Aspects of RKHS relevant to the Koopman Operator}
\begin{tabular}{p{3cm}|p{6cm}|p{5cm}}
\textbf{Aspect} & \textbf{Relevance to Koopman Operator} & \textbf{Mathematical Representation} \\[1ex] \hline 
Function Space & Koopman operator maps functions (observables); RKHS provides structured function spaces. & $\mathcal{F} \subseteq \mathcal{H}, \quad \mathcal{K}^t: \mathcal{F} \to \mathcal{F}$ \\[1ex]
 & & \\[0.2ex]
Discrete Data & Handles point evaluations of functions, key for real-world data-driven analysis. & $\boldsymbol{\psi}(\omega) = \langle \boldsymbol{\psi}, k(\cdot, \omega) \rangle$ \newline (reproducing property) \\[1ex] 
 & & \\[0.2ex]
Kernel Function & Efficiently constructs feature spaces for function representation. & $k(\omega, \omega') = \langle \boldsymbol{\psi}(\omega), \boldsymbol{\psi}(\omega') \rangle_\mathcal{H}$ \\[1ex] 
 & & \\[0.2ex]
Product Spaces & Represents multi-dimensional interactions between variables. & $\mathcal{H}_{\text{product}} = \mathcal{H}_E \otimes \mathcal{H}_C$ \\[1ex] 
 & & \\[0.2ex]
Theoretical \newline Simplicity & Ensures boundedness, completeness, and well-defined operator properties. & $\| \mathcal{K}\boldsymbol{\psi} \| \leq C \|\boldsymbol{\psi}\|$ \newline (bounded linear operator) \\[1ex] 
 & & \\[0.2ex]
Data-Driven \newline Approximations & Enables finite-dimensional approximations for computational feasibility. & $\mathcal{K}^t \approx K^t := \Psi^t \Psi^\dagger$ \newline (this paper) \\ 
\end{tabular}
\label{tab:rkhs_aspects}
\end{table}


\newpage
\section{\texttt{Kausal}: A Differentiable Deep Koopman Causal Analysis Module}
\label{si-sec:kausal_api}
In this section, we highlight code snippets applying the \texttt{Kausal} algorithm. Full API documentation and open-source code is available at \url{https://github.com/juannat7/kausal}.

\textbf{Setting up}. First, you can define multivariate timeseries and assign them as cause and effect variables by which a causal analysis is going to be conducted. 

\begin{figure}[h]
\begin{lstlisting}[language=Python]
import torch
from kausal.koopman import Kausal

# Define your cause/effect variables
cause = torch.tensor(...)
effect = torch.tensor(...)

# Initialize `Kausal` object
model = Kausal(cause = cause, effect = effect)
\end{lstlisting}
\end{figure}

\textbf{Performing causal analysis}. You can then perform causal analysis by providing \texttt{time\_shift} parameter (defaults: 1).

\begin{figure}[h]
\begin{lstlisting}[language=Python]
# Estimate causal effect
causal_effect = model.evaluate(time_shift = 1)
\end{lstlisting}
\end{figure}

\textbf{Defining custom observable functions}. You can then specify specific observable functions, e.g., \texttt{MLP} or \texttt{CNN}. By default, it will use \texttt{RFF} with dictionary of size $M=500$. If you choose a learnable dictionary, you can fit them given hyperparameters.

\begin{figure}[h!]
\begin{lstlisting}[language=Python]
from kausal.observables import MLPFeatures

model = Kausal(
    marginal_observable = MLPFeatures(...), 
    joint_observable = MLPFeatures(...), 
    cause = torch.tensor(...),
    effect = torch.tensor(...)
)

# Fit both marginal and joint observables if using learnable dictionaries.
marginal_loss, joint_loss = model.fit(
    n_train = n_train, 
    epochs = epochs, 
    lr = lr, 
    batch_size = n_train,
    **kwargs
)
\end{lstlisting}
\end{figure}

\textbf{Changing decomposition method}. Several regression techniques to estimate the Koopman operator are also provided, e.g., full-rank pseudo-inverse (\texttt{pinv}) or low-rank mode decomposition (\texttt{DMD}).

\begin{figure}[h!]
\begin{lstlisting}[language=Python]
from kausal.regressors import DMD

model = Kausal(
    regressor = DMD(svd_rank = 4),
    cause = torch.tensor(...),
    effect = torch.tensor(...)
)
\end{lstlisting}
\end{figure}

\newpage
\section{Experiments}
\label{si-sec:experiments}
\subsection{Coupled R\"{o}ssler oscillation}
\label{si-sec:experiments_coupled_rossler}

\begin{figure}[ht!]
    \centering
    \includegraphics[width=1\linewidth]{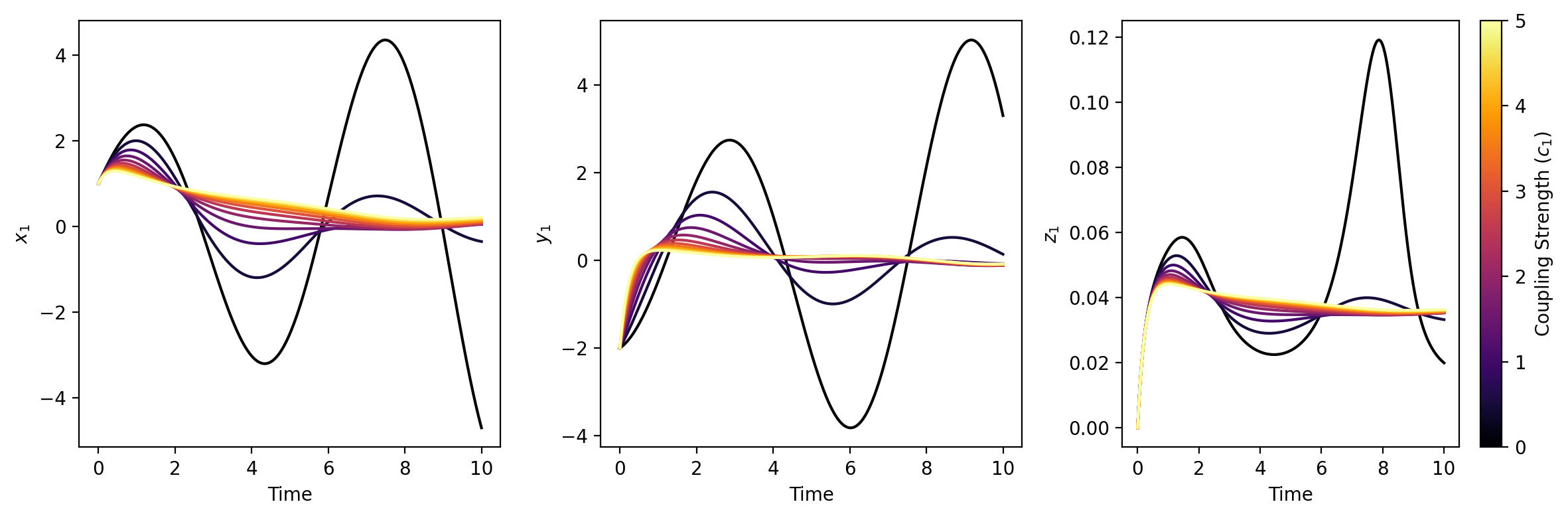}
    \caption{Coupled R\"{o}ssler Oscillation with varying coupling strength $c_1$, when $c_2 = 0$.}
    \label{si-fig:coupled_rossler}
\end{figure}

\textbf{Experimental setup.} The system is solved with Dormand Prince 5th-Order scheme with $t \in [0, 10]$ and $\Delta t = 10^{-2}$. The parameters used here are $a = 0.2, b = 0.2, d = 5.7, \varphi_1 = 1.0, \varphi_2 = 1.0, c_1 = 0.5, c_2 = 0.0$. In the case of the dictionary learning setting, we parameterize the lifting functions $\boldsymbol{\psi}_\theta$ of the marginal and joint models with a 2-layer Multi-Layer Perceptron (MLP) with hidden channels of $[16, 32]$, activated by \texttt{sigmoid}, and optimized with \texttt{AdamW} using a learning rate of $10^{-2}$ over 500 epochs. Otherwise, in the prescribed dictionary setting, we use Random Fourier Feature (RFF) of size $M=500$ as the default. Figure~\ref{si-fig:coupled_rossler} illustrates a sample trajectory when varying coupling strength $c_1$ while setting $c_2 = 0$.

We also perform similar conditional forecasting experiments as in Figure~\ref{fig:coupled_rossler_forecasting}, but using RFF to estimate observables.

\begin{figure}[h!]
    \centering
    \begin{subfigure}[b]{0.48\textwidth}
        \centering
        \includegraphics[width=\textwidth]{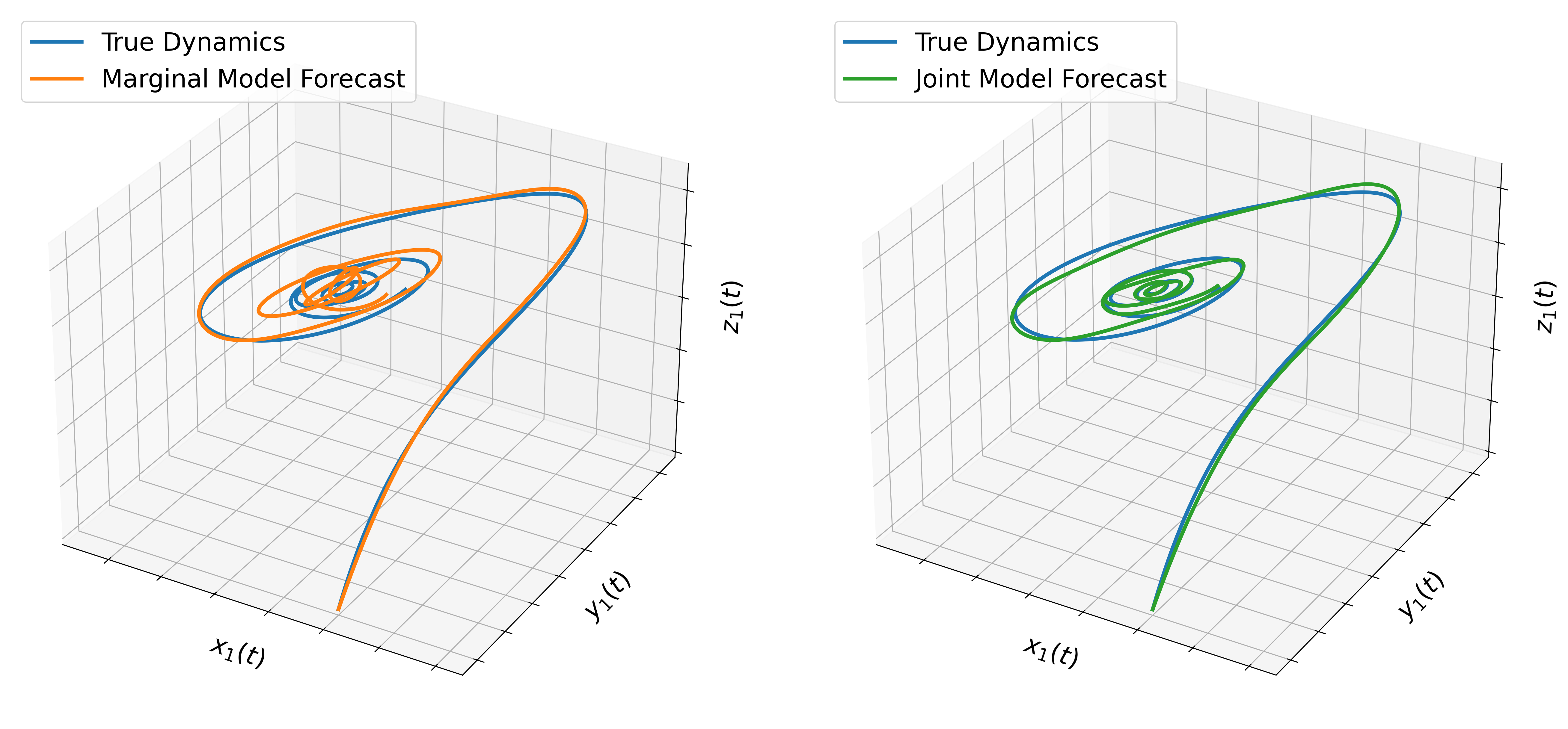}
        \caption{Conditional forecasts in the \textit{true} causal direction $\Omega_C \rightarrow^t_K \Omega_E$}
    \end{subfigure}
    \hfill
    \begin{subfigure}[b]{0.48\textwidth}
        \centering
        \includegraphics[width=\textwidth]{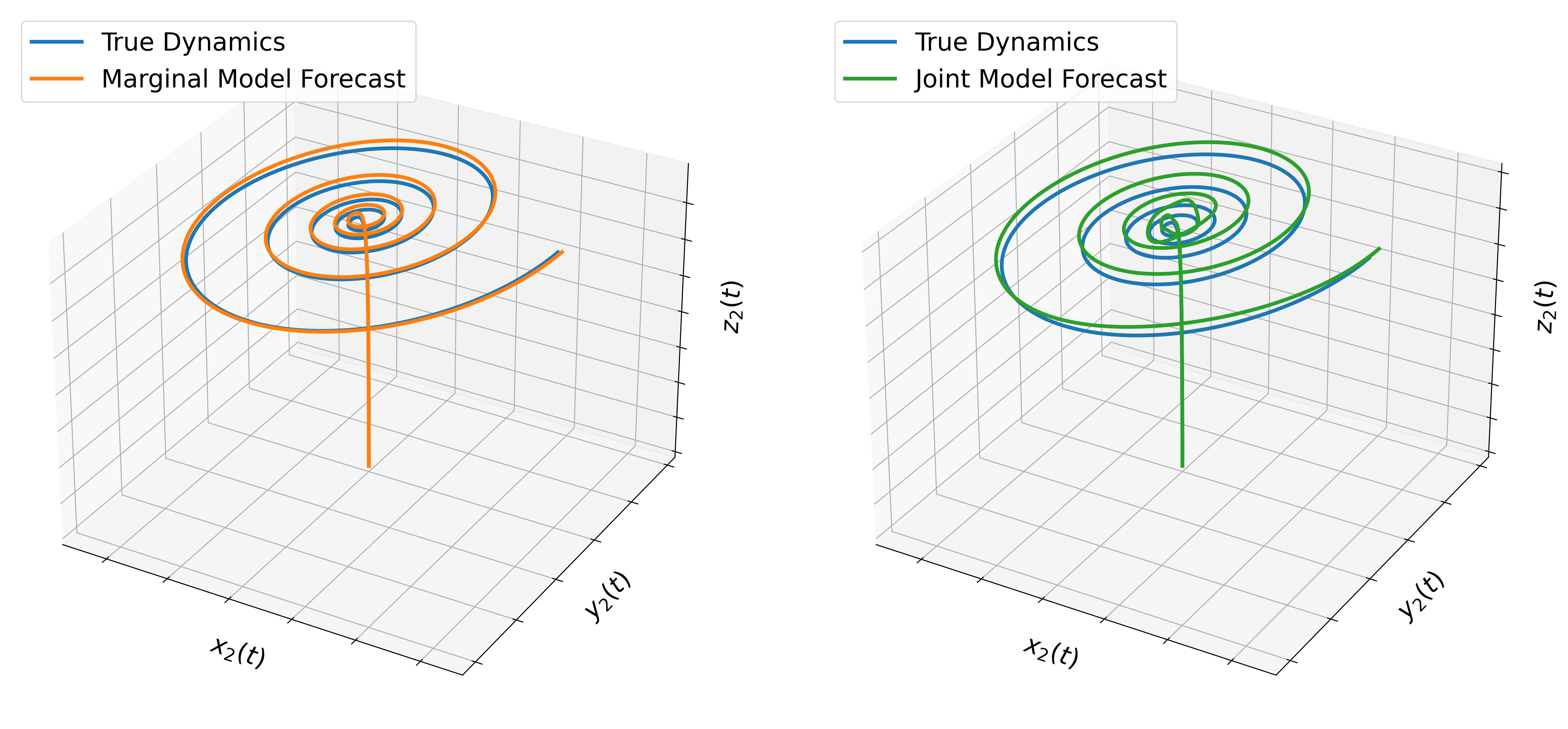}
        \caption{Conditional forecasts in the \textit{false} causal direction $\Omega_E \rightarrow^t_K \Omega_C$}
    \end{subfigure}
    \caption{Conditional forecasts in the (a) \textit{true} and (b) \textit{false} causal direction using an RFF kernel. In (a), the addition of $\Omega_C$ in the joint model improves the forecast of $\Omega_E$ relative to the the marginal model where it is excluded. However, in (b), the addition or exclusion of $\Omega_E$ has no effect on the forecast of $\Omega_C$. Nonetheless, the qualitative separations are not as distinctive as those using MLP-based observables.}
    \label{si-fig:coupled_rossler_forecasting}
\end{figure}

\subsection{Reaction-diffusion equation}
\label{si-sec:experiments_reaction_diffusion}
\textbf{Experimental setup.} The system is solved over a $16 \times 16$ grid with Dormand Prince 5th-Order scheme with $t \in [0, 10]$ and $\Delta t = 10^3$. The parameters used here are $D_u = D_v = 0.1$ as the diffusion coefficients, $a = b = 0.3$ as the reaction parameters. As described in the main text, the coupling terms are set as $\beta = 10$ and $\gamma = 0$. In the case of the dictionary learning setting, we parameterize the lifting functions $\boldsymbol{\psi}_\theta$ of the marginal and joint models with a convolution-based encoder-decoder symmetric structure (CNN) with hidden channels of $[16, 32, 64, 128]$, activated by $sigmoid$, and optimized with $AdamW$ using a learning rate of $10^{-4}$ over 50 epochs. Otherwise, in the prescribed dictionary setting, we use Random Fourier Feature (RFF) of size $M=500$ as the default.  

\begin{figure}[ht!]
    \centering
    \includegraphics[width=1\linewidth]{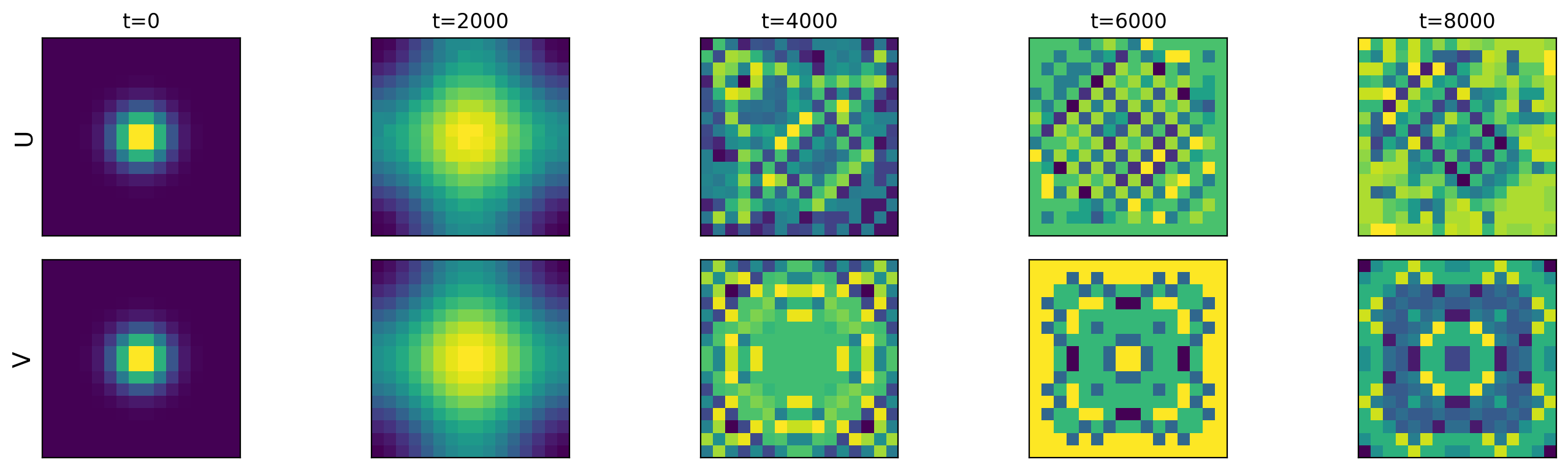}
    \caption{Nonlinear and coupled reaction-diffusion equation sample realization for different time steps.}
    \label{si-fig:reaction_diffusion}
\end{figure}

\subsection{El Niño–Southern Oscillation}
\label{si-sec:experiments_enso}
We analyze both simulated and real-world datasets of El Niño–Southern Oscillation (ENSO). We first describe the simulation setup before elaborating on the experiments.

\textbf{Experimental setup.} The system is solved with Dormand Prince 5th-Order scheme with $t \in [0, 100]$ and $\Delta t = 10^{-2}$. The parameters used here are $r = 0.25, \alpha = 0.125, b_0 = 2.5, c = 1.0, \gamma = 0.75$. As described in the main text, we vary both the coupling term $\mu$ and nonlinearity parameter $\epsilon$. In the case of the dictionary learning setting, we parameterize the lifting functions $\boldsymbol{\psi}_\theta$ of the marginal and joint models with a 2-layer Multi-Layer Perceptron (MLP) with hidden channels of $[16, 32]$, activated by \texttt{sigmoid}, and optimized with \texttt{AdamW} using a learning rate of $10^{-2}$ over 500 epochs. Otherwise, in the prescribed dictionary setting, we use Random Fourier Feature (RFF) of size $M=500$ as the default. 

\begin{figure}[h!]
    \centering
    \begin{subfigure}[b]{0.4\textwidth}
        \centering
        \includegraphics[width=\textwidth]{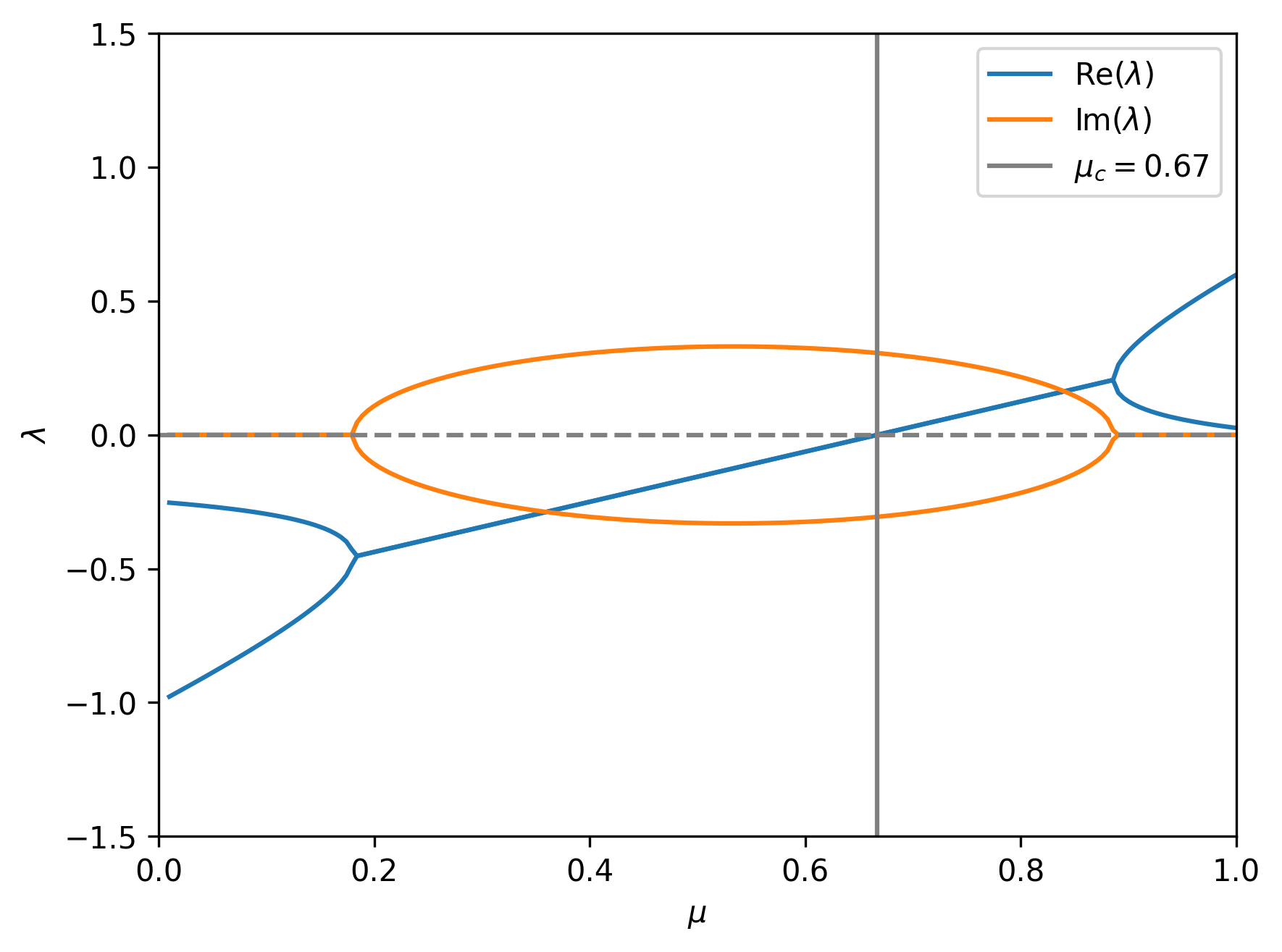}
        \caption{Bifurcation diagram with varying $\mu$. The critical point is reached when $\mu_c = 0.67$. Meanwhile, $\mu < \mu_c$ and $\mu > \mu_c$ represent dissipative and chaotic dynamics respectively. }
    \end{subfigure}
    \hfill
    \begin{subfigure}[b]{0.55\textwidth}
        \centering
        \includegraphics[width=\textwidth]{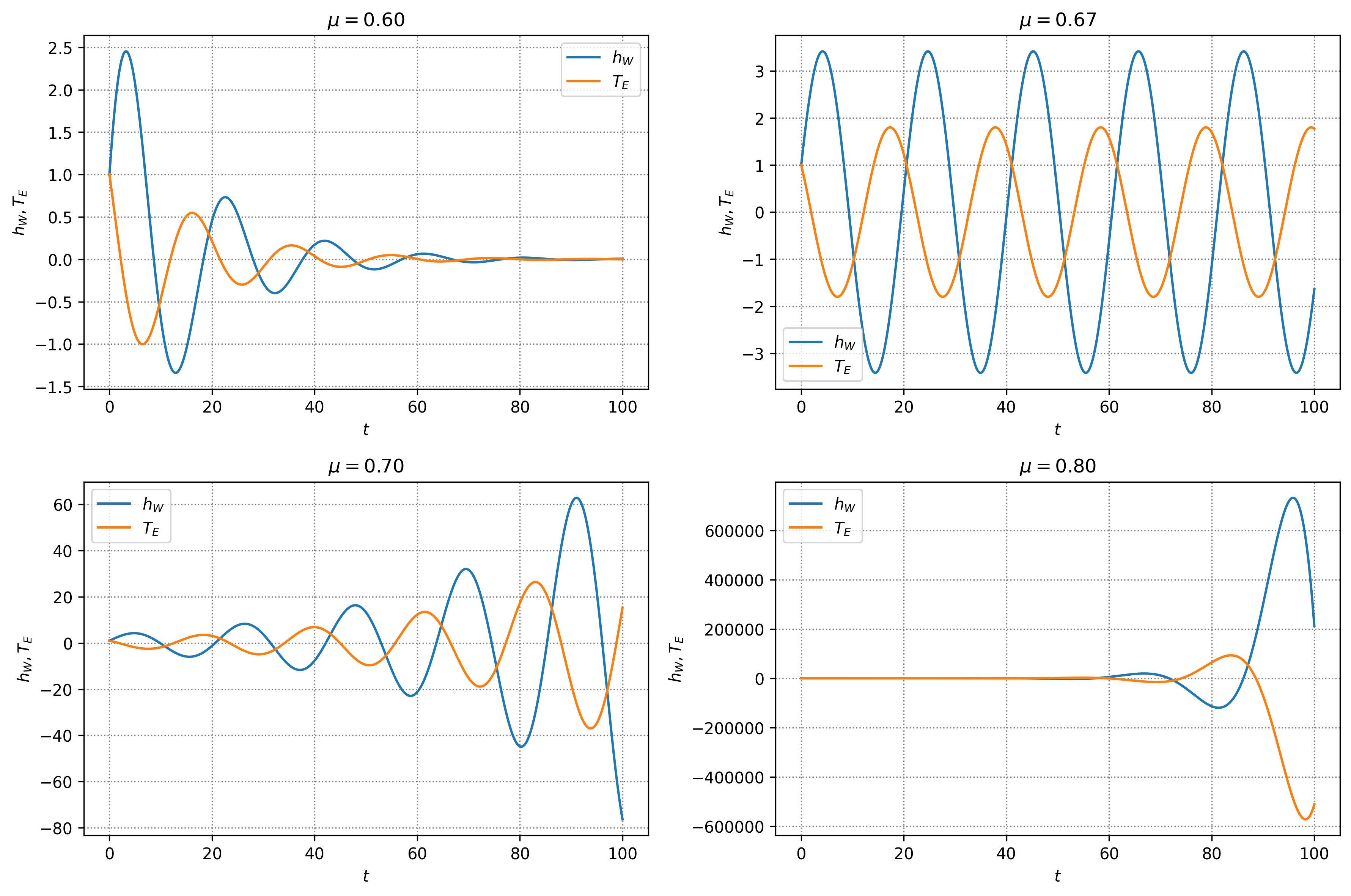}
        \caption{Sample trajectories given identical initial condition but different coupling terms $\mu$. We only analyze the coupled oscillatory regime because the decoupled mechanism is physically unrealistic (i.e., $\text{Im}(\lambda) \neq 0$).}
    \end{subfigure}
    \caption{Linear stability analysis of ENSO model by \citet{jinEquatorialOceanRecharge1997} showcasing in a) the bifurcation diagram that identifies fixed point structures, and in b) sample trajectories given different coupling strength $\mu$ that produce dissipative, stable, or chaotic oscillatory realizations (i.e., $\text{Im}(\lambda) \neq 0)$.}
    \label{si-fig:enso_linear_analysis}
\end{figure}

\textbf{Linear stability analysis.} Here, we provide more details on our linear stability analysis, including a bifurcation diagram and sample trajectories around the critical point $\mu_c = 0.67$ (see Figure~\ref{si-fig:enso_linear_analysis}a).

\begin{figure}[h!]
    \centering
    \includegraphics[width=0.5\linewidth]{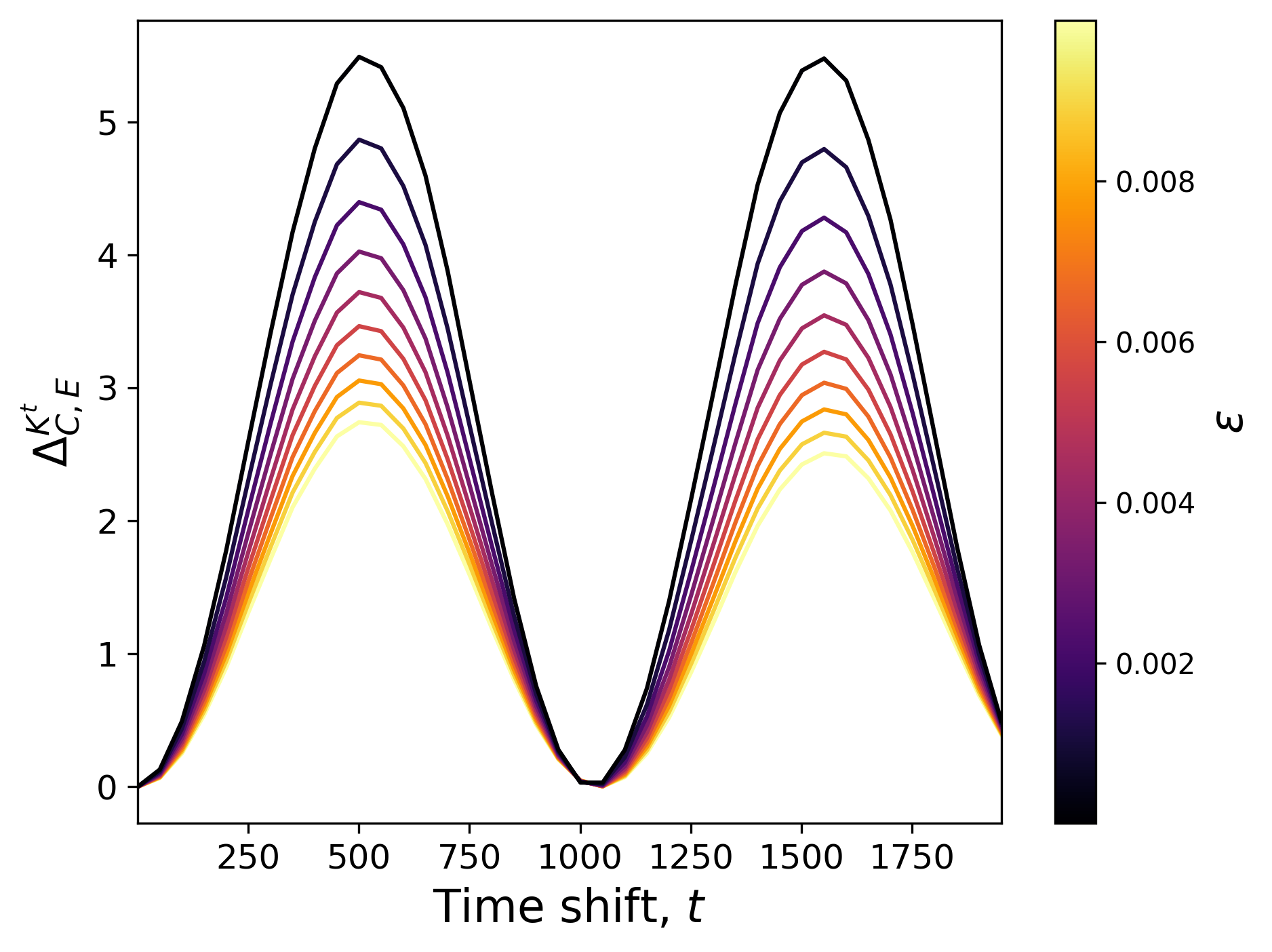}
    \caption{Causal measure of ENSO by varying nonlinearity strength $\epsilon$ using RFF kernels to approximate the observables.}
    \label{si-fig:enso_estimation_rff}
\end{figure}

\textbf{Capturing nonlinearity strength.} In Figure~\ref{si-fig:enso_estimation_rff}, we perform causal measure estimations by varying $\epsilon$ similar to Figure~\ref{si-fig:enso_estimation_rff} but using RFF kernels ($M=500$). 

\textbf{Application to real-world data.} Lastly, we move beyond simulation and to a real-world application, using a long-term ocean reanalysis product. 

\begin{figure}[h!]
    \centering
    \includegraphics[width=0.8\linewidth]{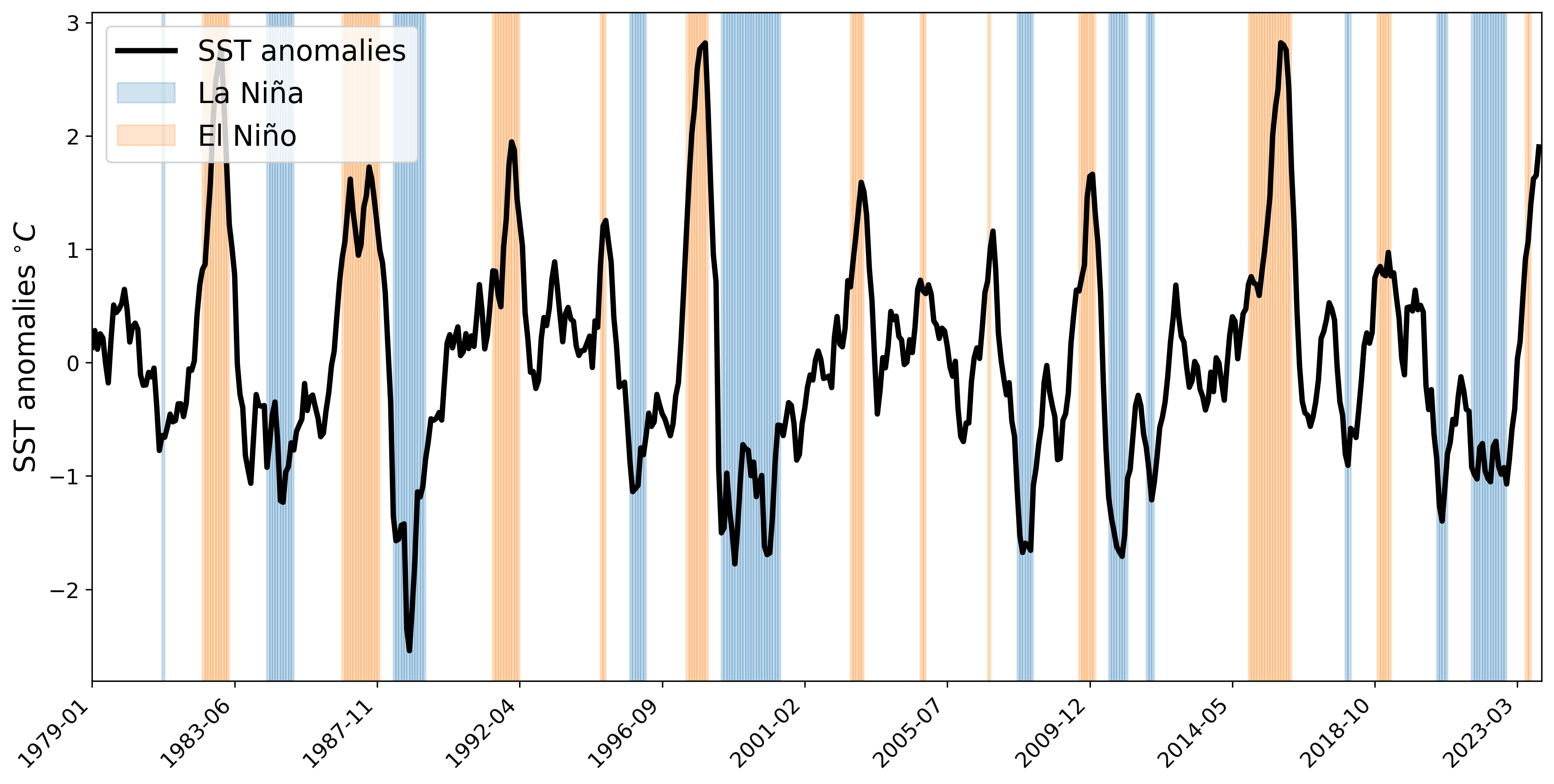}
    \caption{Sea surface temperature (SST) anomalies (black line) in Nino3.4 region to identify major El Ni\~no and La Ni\~na events (shading) according to NOAA's ONI index \cite{glantz2020OniIndex}.}
    \label{si-fig:sst_anomalies}
\end{figure}

As mentioned briefly in the main text, we use the ECMWF monthly $1.5^\circ$ ocean reanalysis product \cite{zuo2019ecmwf} as preprocessed in \citet{nathaniel2024chaosbench}. In particular, we use anomalies of \texttt{sosstsst} (sea surface temperature) for $T$ and \texttt{sossheig} (sea surface height) for $h$ in Equation~\ref{eq:enso}. Note, that while in the ENSO model described above $h$ resembles the thermocline height, we use sea surface height in this observation-based study. Both variables are interchangeable when studying the feedback mechanisms of ENSO~\citep[as the sea level height decreases the thermocline shallows;][]{zhao2021breakdown}. We define the equatorial band as $5^\circ$S-$5^\circ$N and then estimate $T$ over the Nino-3 region ($150^\circ$E-$90^\circ$W) and $h$ over the Nino-4 region ($150^\circ$E-$150^\circ$W). The background SST anomaly is computed along the Nino3.4 region ($170^\circ$W-$120^\circ$W). Lastly, we deseasonalize the variables by subtracting the 30-year monthly climatology. Following NOAA's Oceanic Ni\~no Index \citep[ONI;][]{glantz2020OniIndex}, El Ni\~no events are defined when at least five consecutive 3-month running means of SST anomalies in the Nino3.4 region show $T \geq 0.5^\circ\text{C}$, and La Ni\~na events when $T \leq -0.5^\circ\text{C}$ (see Figure~\ref{si-fig:sst_anomalies}).

\begin{figure}[h!]
    \centering
    \includegraphics[width=0.8\linewidth]{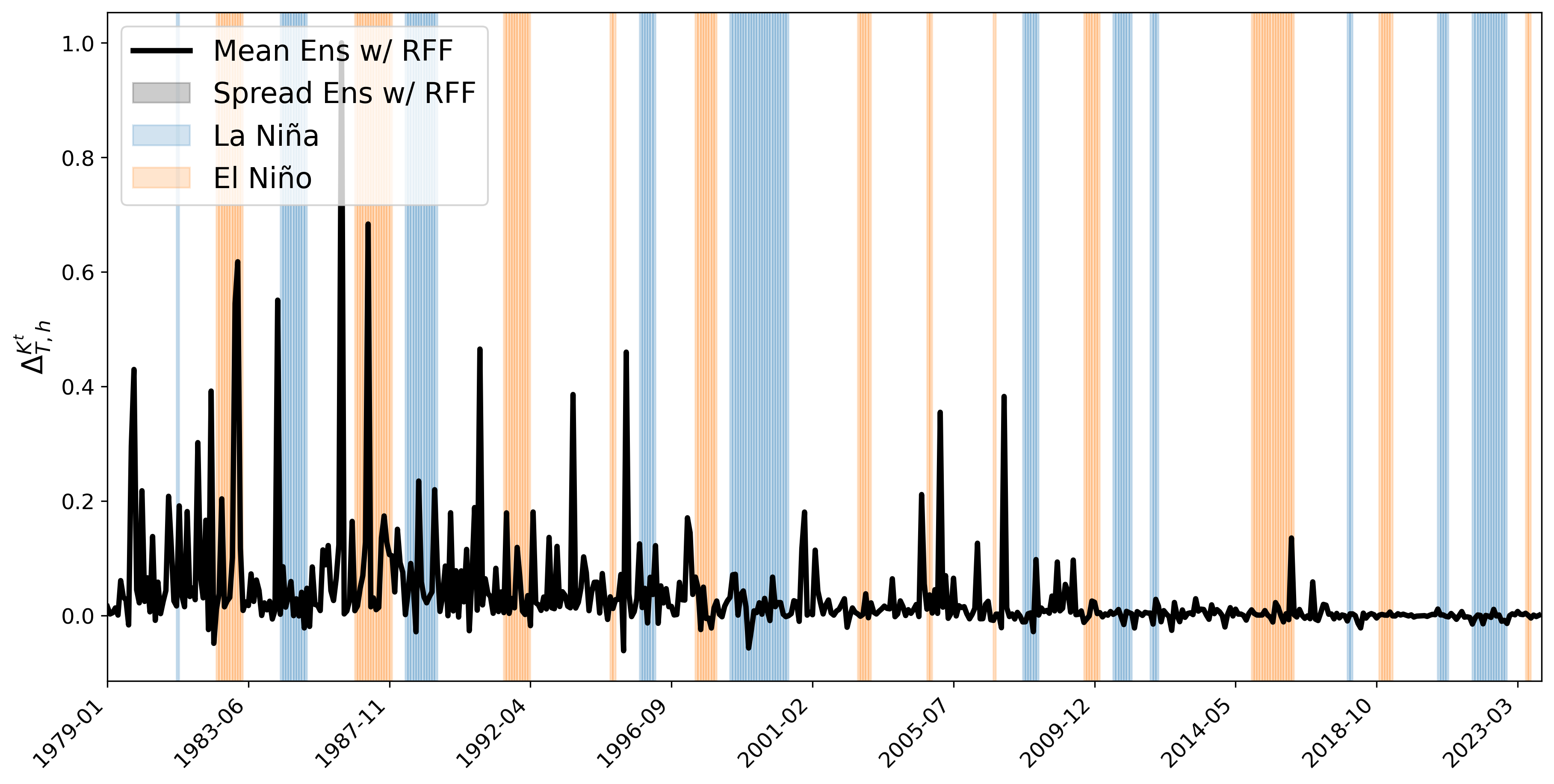}
    \caption{Causal measure of ENSO using real-world data showcasing the use of RFF-derived observables results in failure to detect meaningful causal signals around major ENSO events.}
    \label{si-fig:enso_real_rff_expanded}
\end{figure}

Instead of computing an area average, we use each gridcell as its own independent dimensionality to increase the number of features. The marginal and joint models are estimated using MLP kernels with 2 hidden-channels of size [512, 1024], activated by \texttt{sigmoid}, and optimized with \texttt{AdamW} using a learning rate of $10^{-2}$ over 500 epochs. Causal measures for the displayed time periods in Figure~\ref{fig:enso_real_mlp} and Figure~\ref{si-fig:enso_real_rff_expanded} are estimated by iteratively increasing time shifts $t$ in Equation~\ref{eq:causal_measure_loss} for fixed initial conditions $\omega_0$, which is given by the respective first displayed date.


In Figure~\ref{si-fig:enso_real_rff_expanded}, an additional analysis of the observed ENSO highlights the limitations of prescribed observables through the failure of RFF to capture causal signals in major ENSO events .

\newpage
\section{Ablation}\label{si-sec:ablation}
We run additional ablation studies to benchmark \texttt{Kausal} with state-of-the-art causal discovery for dynamical systems.

\subsection{Task 1: Causal direction identification}\label{app:task1}
We perform hypothesis testing to identify $\Omega_C \rightarrow \Omega_E$. For this work, we use the Coupled R\"{o}ssler oscillators and reaction-diffusion equation where the true directionality is known and clearly distinguishable (no feedback mechanism i.e., $\Omega_C \rightarrow \Omega_E$ and $\Omega_E \not \rightarrow \Omega_C$). For \texttt{Kausal}, we propose a statistical test to check $\Omega_C \rightarrow \Omega_E$, given by Algorithm \ref{alg:hypo_test}, drawing inspiration from \citet{christiansen2022toward}. In plain English, the algorithm checks if there is distinguishable causal effect in the direction of $\Omega_C \rightarrow \Omega_E$ when compared with an $N$-times randomly shuffled timeseries for a given time shift $t$. By default, we use $N=100$ and $t=1$ and the significance is measured with a one-sided test.

\begin{algorithm}[h]
   \caption{\texttt{Kausal} identifiability test ($\Omega_C \rightarrow \Omega_E$)}
   \label{alg:hypo_test}
\begin{algorithmic}
   \STATE {\bfseries input:} $\Omega_C$, $\Omega_E$, $N$, $t$
   \STATE {\bfseries require:} $N > 1$ \hfill Number of random permutation
   \STATE $\Delta_{C \rightarrow E} \gets \texttt{Kausal}\Bigl(\Omega_C \xrightarrow{t}_K \Omega_E\Bigr)$ \hfill Apply Equation~\ref{eq:causal_measure}
   \STATE $\Delta_{C \not \rightarrow E} \gets \emptyset$ \hfill Initialize causal measure placeholder for randomized timeseries
   \FOR{$n \gets 1$ to $N$} 
   \STATE $\hat{\Omega}_C \gets \texttt{RandomTemporalPermute}(\Omega_C)$
   \STATE $\hat{\Omega}_E \gets \texttt{RandomTemporalPermute}(\Omega_E)$
   \STATE $\Delta_{C \not \rightarrow E}[n] \gets \texttt{Kausal}\Bigl(\hat{\Omega}_C \xrightarrow{t}_K \hat{\Omega}_E\Bigr)$ \hfill Apply Equation~\ref{eq:causal_measure}
   \ENDFOR
   \STATE $\rho \gets$ \texttt{PTest}($\Delta_{C \rightarrow E} > \Delta_{C \not \rightarrow E}$) \hfill Perform one-sided p-test
   \STATE {\bfseries return: $\rho$}
\end{algorithmic}
\end{algorithm}

We use PCMCI+ \cite{runge2020discovering} and VARLiNGAM \cite{hyvarinen2010estimation} as our baselines to check the true and false causal directions as they provide statistical tests out-of-the-box. The baseline setups are as follow:

\begin{itemize}
    \item PCMCI+: $\tau_{max} = 1$ to recover at most $t-1$ time lag with partial correlation (ParCorr) conditional test,
    \item VARLiNGAM: $\tau_{max} = 1$ to recover at most $t-1$ time lag.
\end{itemize}

\subsection{Task 2: Causal magnitude estimation}\label{app:task2}
Since most baselines (PCMCI+ \cite{runge2020discovering}, VARLiNGAM \cite{hyvarinen2010estimation}, cLSTM \cite{tank2021neural}, and TSCI \cite{butler2024tangent}) aggregate causal measures given a timeseries (i.e., estimated causal graph is assumed to be static), we employ an identical sliding window strategy proposed in \texttt{Tigramite} tutorial for ENSO identification analysis \cite{runge2019inferring}. In short, we define a sliding window of size 10 and step size of 1. For each time instance $t \geq 1$, we extract $\Omega_C(t-1) \rightarrow \Omega_E(t)$ causal effect measure. In this experiment, we compute the absolute causal magnitude estimate (i.e., unsigned) and evaluate it using Area Under the Receiver Operating Characteristic (AUROC) \cite{peterson1954theory} where the binary classification include both positive and negative extrema about a pre-defined $\pm\gamma\sigma$, where $\gamma \in [0,1]$ and $\sigma$ is the deviation about a mean state (e.g., in our real ENSO example, the extrema refer to El Ni\~no and La Ni\~na, and the mean state is defined as the climatology). For toy problems, we prescribe $\gamma_\texttt{Coupled-Rossler} := 1.00$ and $\gamma_\texttt{reaction-diffusion} := 0.50$ to sufficiently capture meaningful extremes. Note that in the reaction-diffusion experiment, we run baselines with a spatially-averaged timeseries as most baseline algorithms do not scale in high-dimensional, multivariate nodes. We use the default hyperparameters settings across models. Finally, we extract the final $L=1000$ time samples in the Coupled R\"{o}ssler oscillations, and further subsample at every $S=50$ steps in the reaction-diffusion trajectory, to mimic how dynamics are typically observed in real-world (i.e., steady-state behaviors and sparse sampling, respectively).

\begin{figure}[h!]
    \centering
    \begin{subfigure}[b]{\textwidth}
        \centering
        \includegraphics[width=\textwidth]{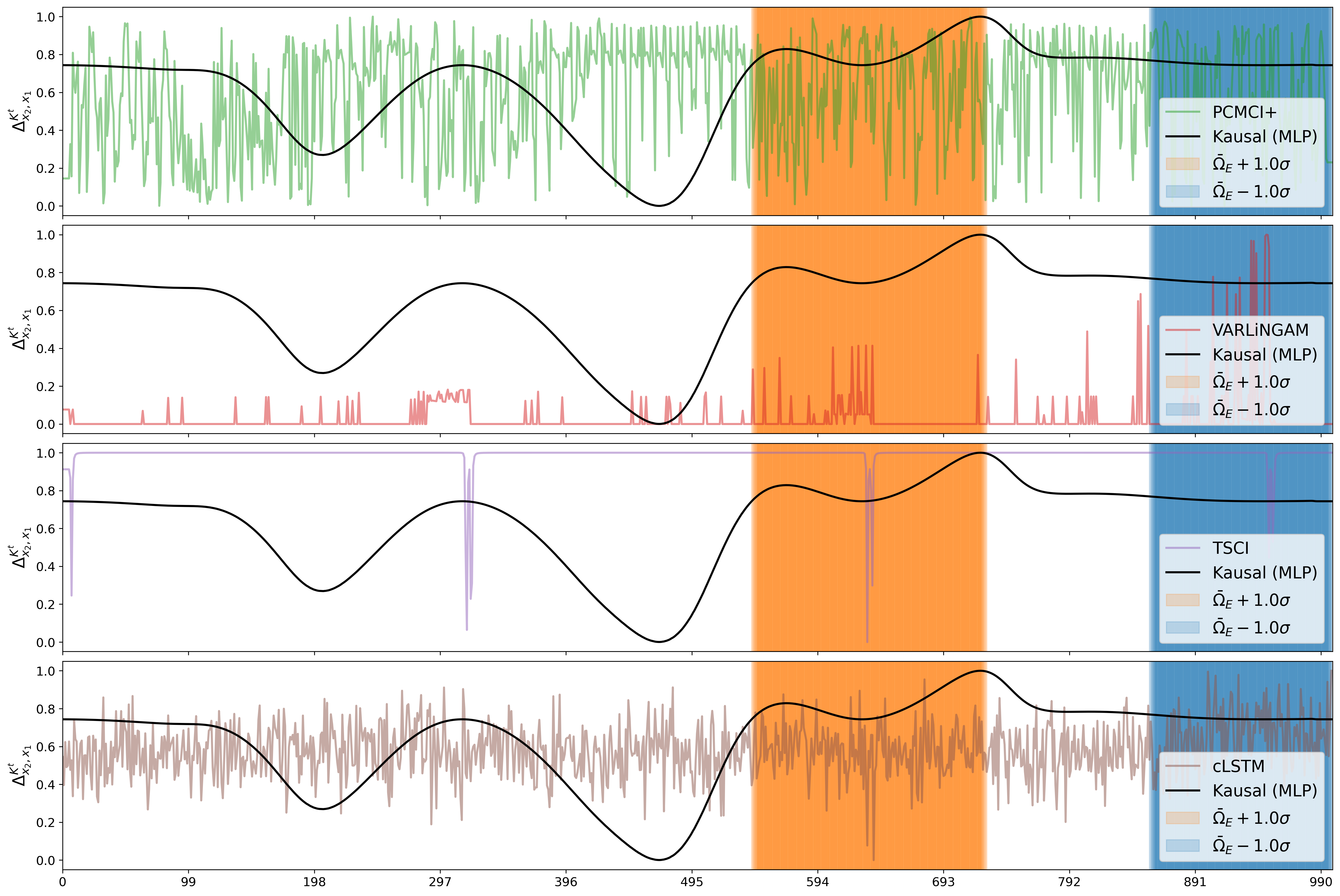}
        \caption{\texttt{Kausal} (MLP) and baselines}
    \end{subfigure}
    \hfill
    \begin{subfigure}[b]{\textwidth}
        \centering
        \includegraphics[width=\textwidth]{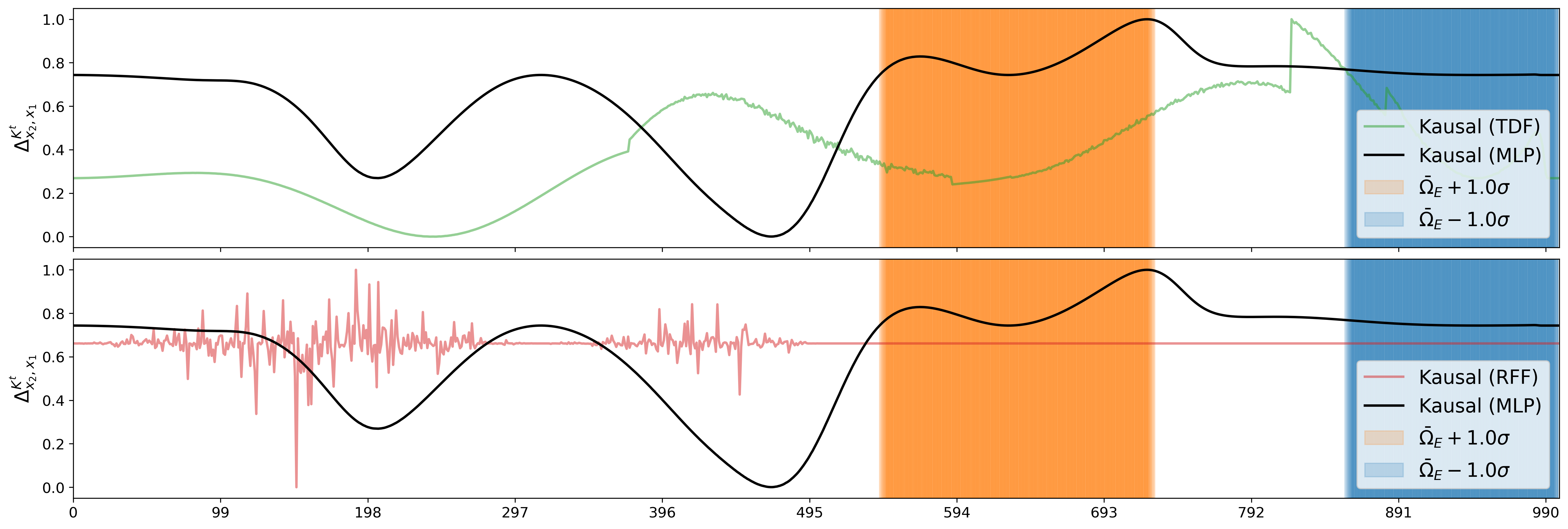}
        \caption{\texttt{Kausal} variants}
    \end{subfigure}
    \caption{Causal magnitude estimation for Coupled R\"{o}ssler oscillators given a backdrop of extremes to be detected. }
    \label{si-fig:coupled_rossler_estimation}
\end{figure}

\begin{figure}[h!]
    \centering
    \begin{subfigure}[b]{\textwidth}
        \centering
        \includegraphics[width=\textwidth]{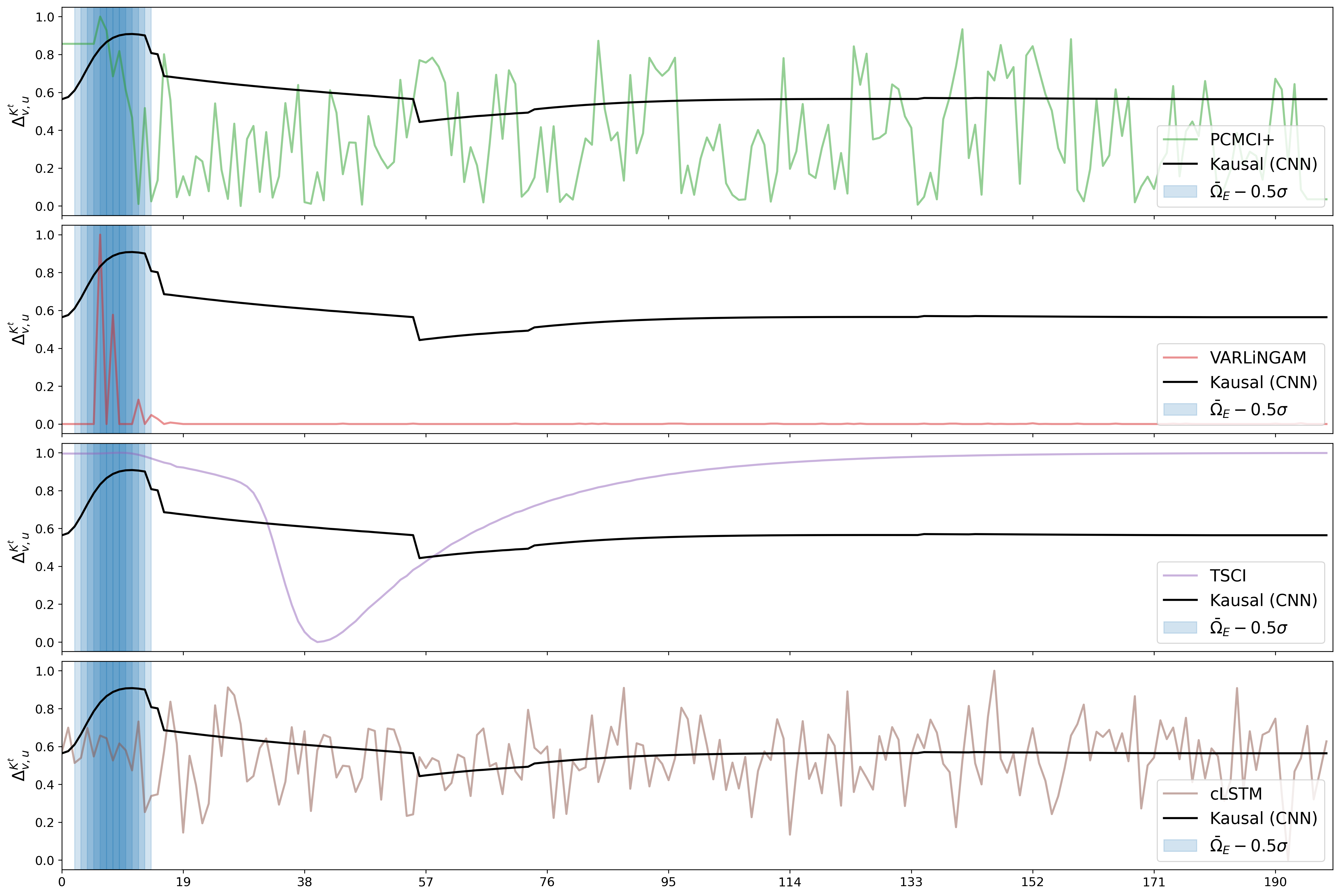}
        \caption{\texttt{Kausal} (CNN) and baselines}
    \end{subfigure}
    \hfill
    \begin{subfigure}[b]{\textwidth}
        \centering
        \includegraphics[width=\textwidth]{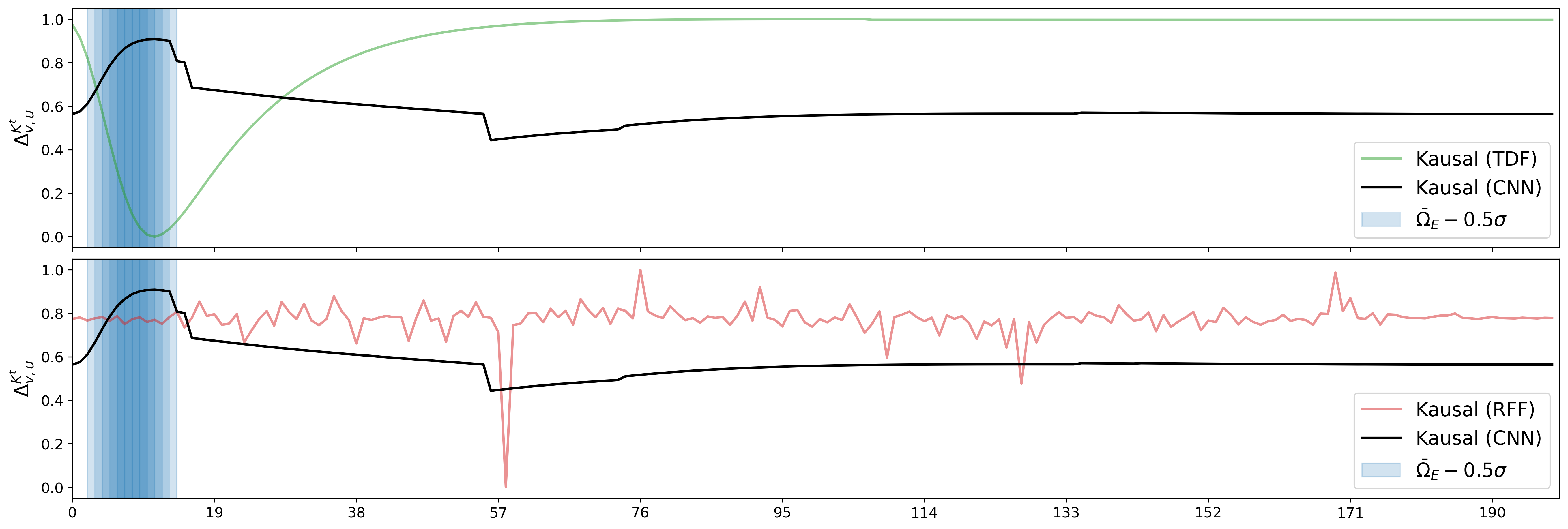}
        \caption{\texttt{Kausal} variants}
    \end{subfigure}
    \caption{Causal magnitude estimation for reaction-diffusion equation given a backdrop of extremes to be detected. }
    \label{si-fig:reaction_diffusion_estimation}
\end{figure}

\begin{figure}[h!]
    \centering
    \begin{subfigure}[b]{\textwidth}
        \centering
        \includegraphics[width=\textwidth]{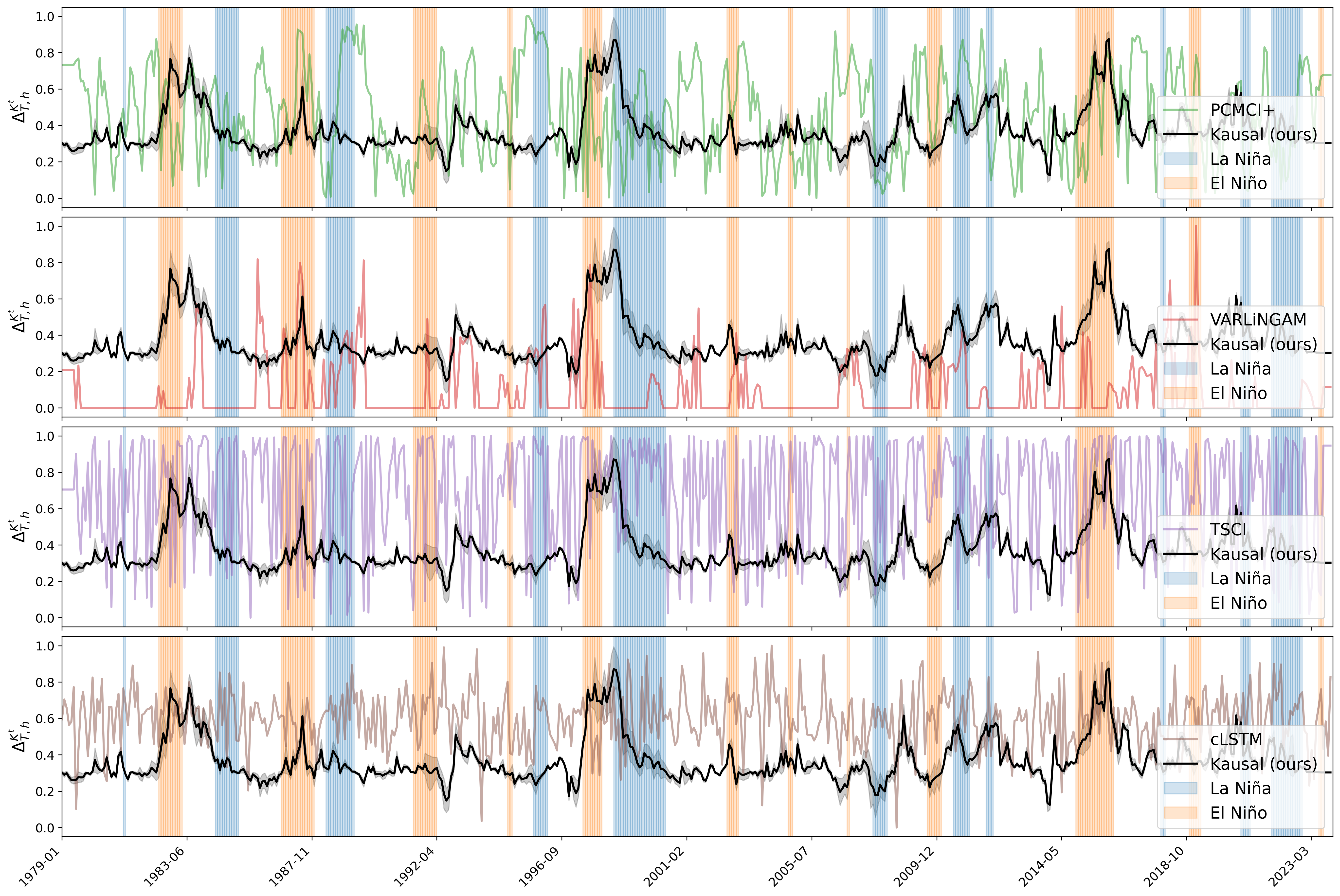}
        \caption{\texttt{Kausal} (MLP) and baselines}
    \end{subfigure}
    \hfill
    \begin{subfigure}[b]{\textwidth}
        \centering
        \includegraphics[width=\textwidth]{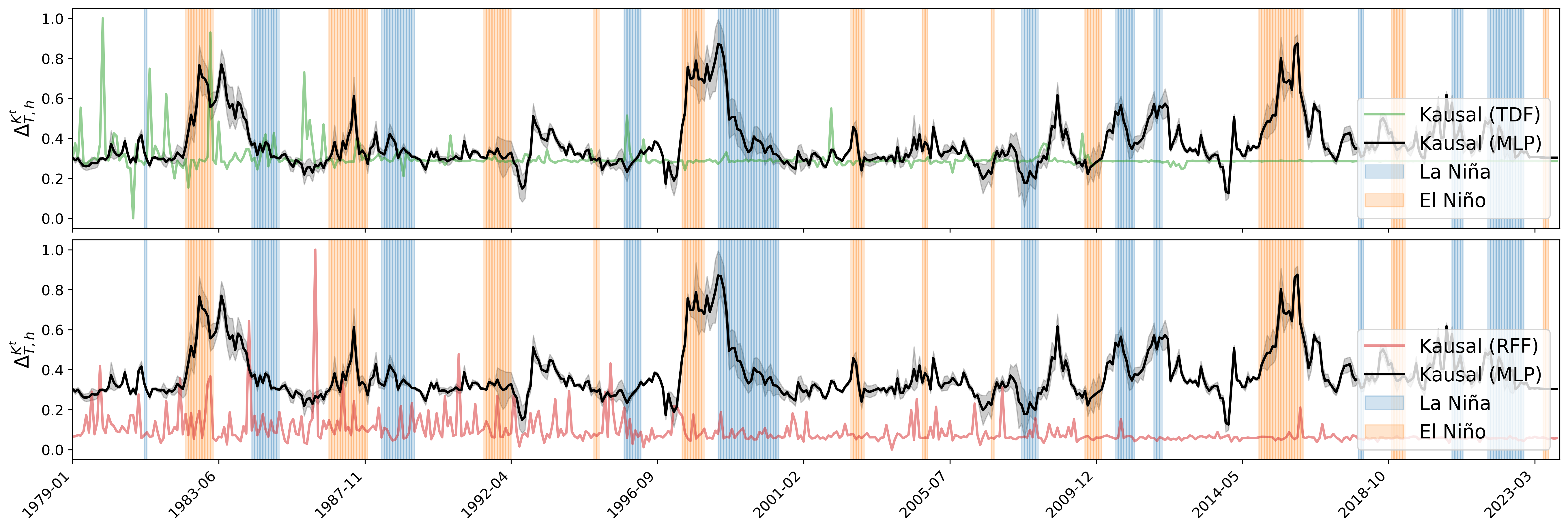}
        \caption{\texttt{Kausal} variants}
    \end{subfigure}
    \caption{Causal magnitude estimation for real ENSO observations given a backdrop of extremes to be detected. }
    \label{si-fig:enso_real_estimation}
\end{figure}

\begin{figure}[h!]
    \centering
    \begin{subfigure}[b]{0.45\textwidth}
        \centering
        \includegraphics[width=0.8\textwidth]{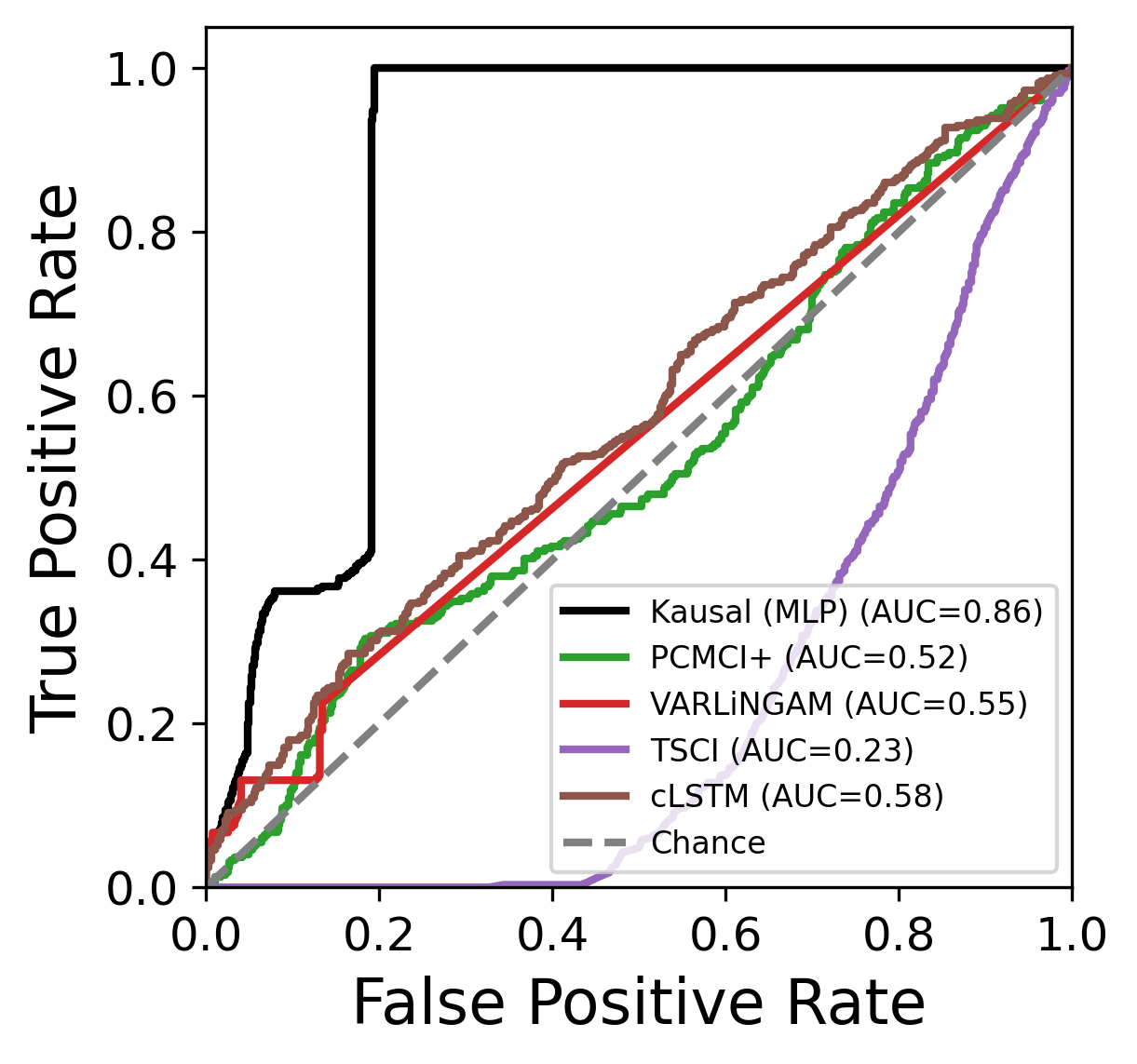}
        \caption{\texttt{Kausal} (MLP) and baselines}
    \end{subfigure}
    \hfill
    \begin{subfigure}[b]{0.45\textwidth}
        \centering
        \includegraphics[width=0.8\textwidth]{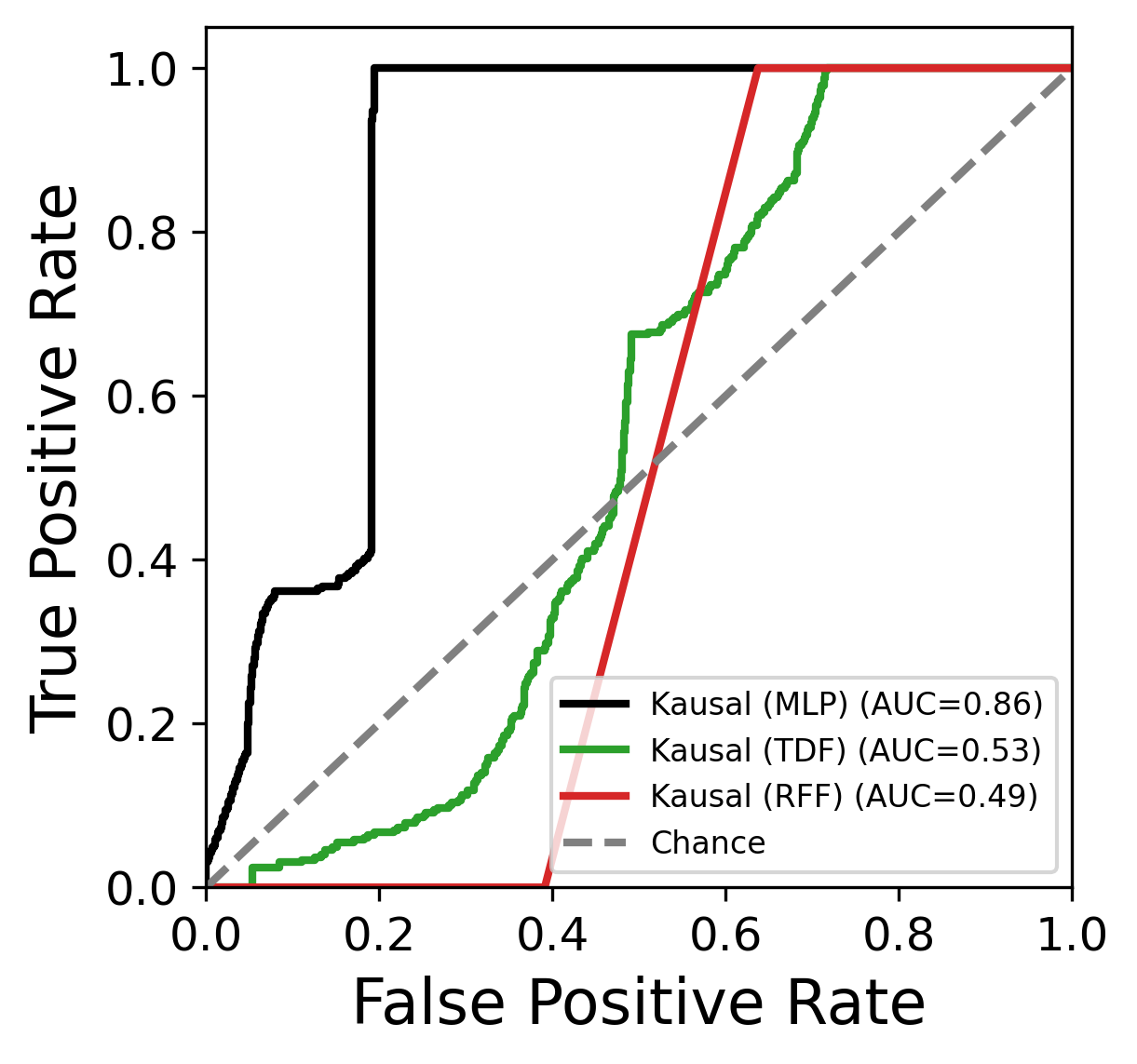}
        \caption{\texttt{Kausal} variants}
    \end{subfigure}
    \caption{AUROC ($\uparrow$ is better) for Coupled R\"{o}ssler oscillators in the causal detection of extreme signals.}
    \label{si-fig:coupled_rossler_auroc}
\end{figure}

\begin{figure}[h!]
    \centering
    \begin{subfigure}[b]{0.45\textwidth}
        \centering
        \includegraphics[width=0.8\textwidth]{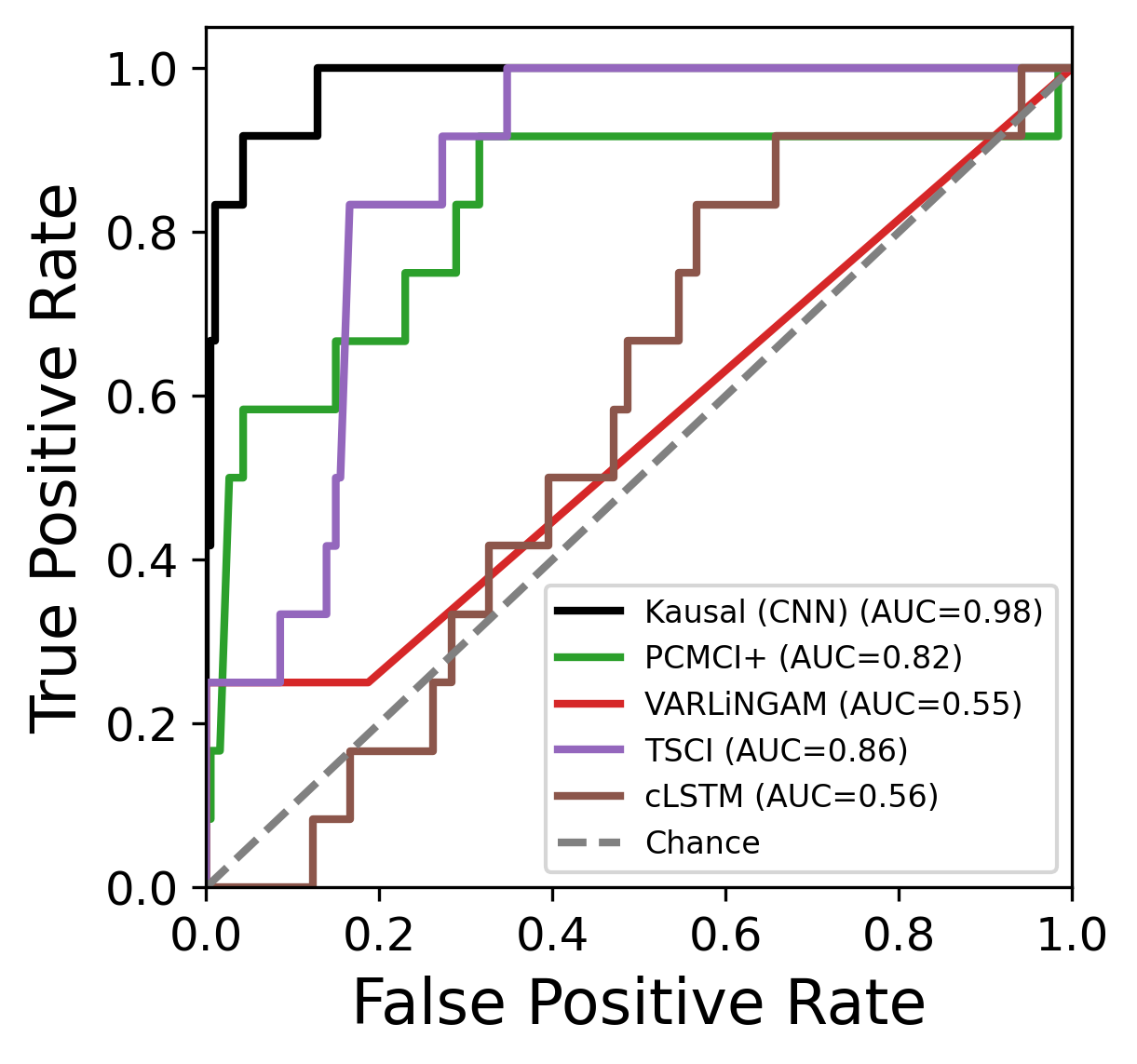}
        \caption{\texttt{Kausal} (CNN) and baselines}
    \end{subfigure}
    \hfill
    \begin{subfigure}[b]{0.45\textwidth}
        \centering
        \includegraphics[width=0.8\textwidth]{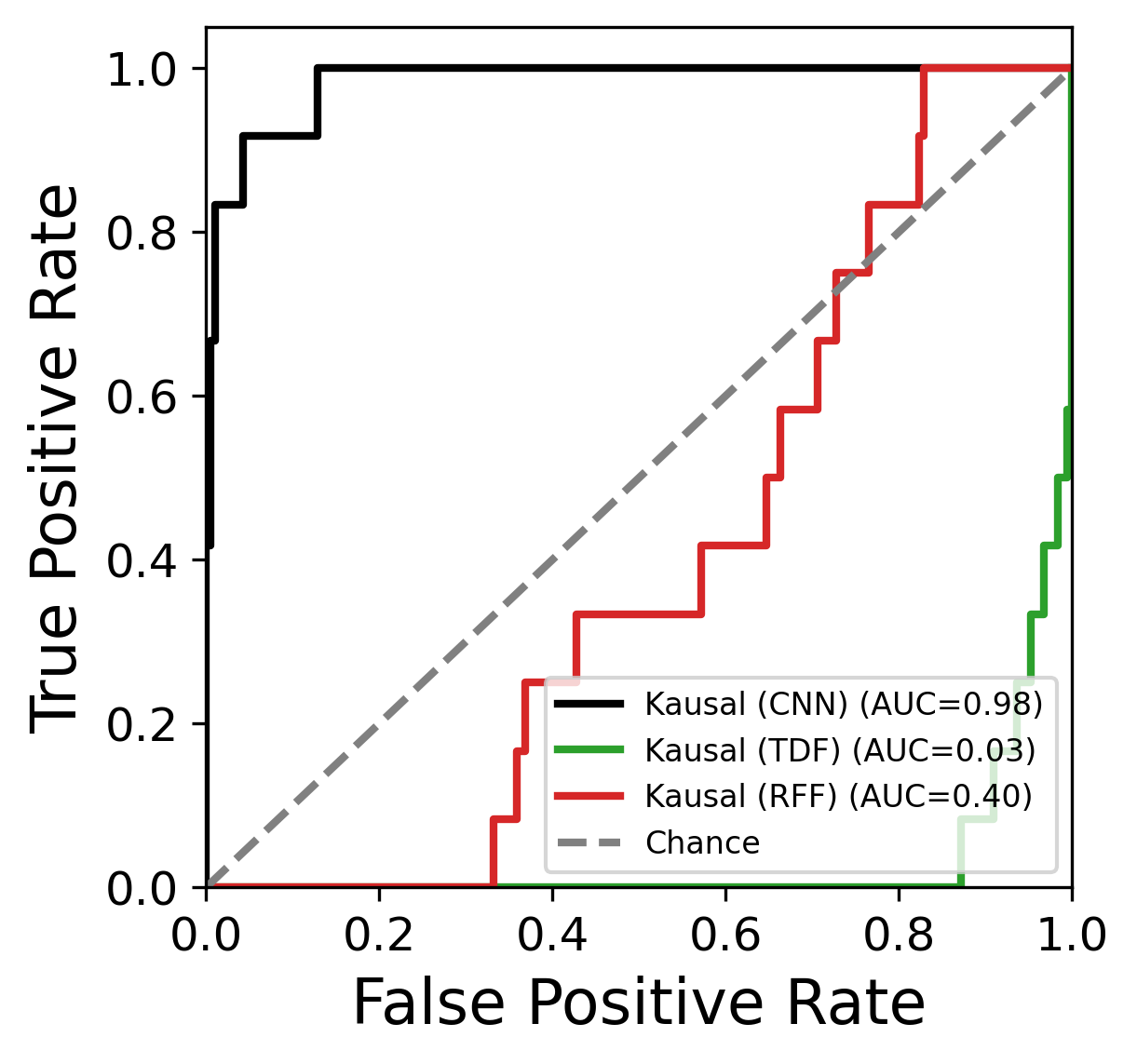}
        \caption{\texttt{Kausal} variants}
    \end{subfigure}
    \caption{AUROC ($\uparrow$ is better) for reaction-diffusion equation in the causal detection of extreme signals.}
    \label{si-fig:reaction_diffusion_auroc}
\end{figure}

\begin{figure}[h!]
    \centering
    \begin{subfigure}[b]{0.45\textwidth}
        \centering
        \includegraphics[width=0.8\textwidth]{docs/enso_real_baselines_auroc.png}
        \caption{\texttt{Kausal} (MLP) and baselines}
    \end{subfigure}
    \hfill
    \begin{subfigure}[b]{0.45\textwidth}
        \centering
        \includegraphics[width=0.8\textwidth]{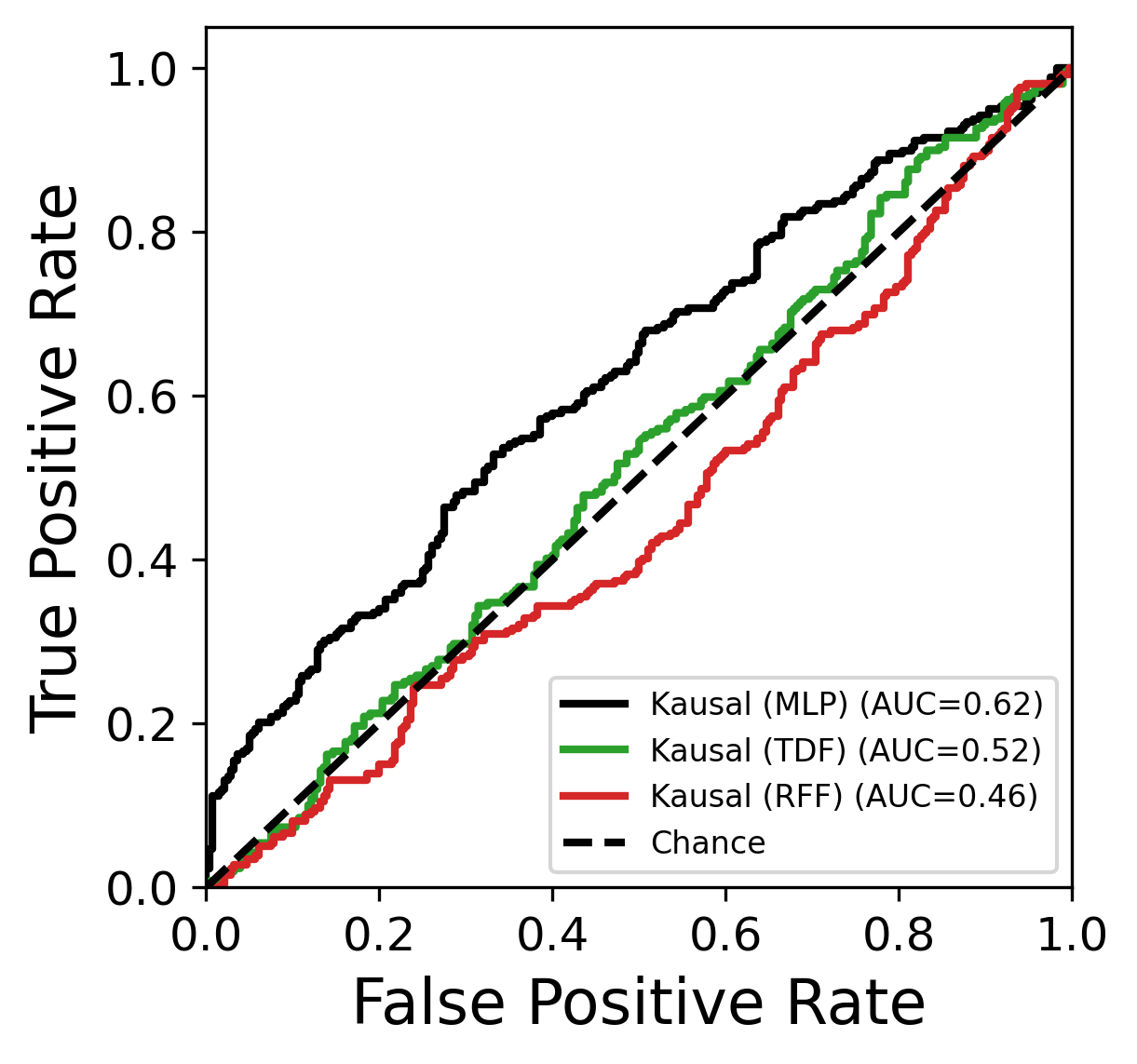}
        \caption{\texttt{Kausal} variants}
    \end{subfigure}
    \caption{AUROC ($\uparrow$ is better) for real ENSO observations in the causal detection of extreme signals.}
    \label{si-fig:enso_real_auroc}
\end{figure}

Figures \ref{si-fig:coupled_rossler_estimation} - \ref{si-fig:enso_real_estimation} illustrate the causal magnitude estimation as timeseries given a backdrop of extremes to be detected. While Figures \ref{si-fig:coupled_rossler_auroc}-\ref{si-fig:enso_real_auroc} highlight the AUROC curve and scores in the causal detection of extreme signals. Across experiments, we find the superiority of \texttt{Kausal} in detecting extreme signals, even in the most challenging real-world ENSO dynamics.

\end{document}